\documentclass[a4]{article}
\usepackage{geometry}
\usepackage{lineno,hyperref}
\usepackage[cmex10]{amsmath}
\usepackage{amssymb}
\usepackage{amsthm}
\usepackage{amsfonts}
\usepackage{algorithmic,algorithm}
\usepackage{tikz}
\usetikzlibrary{bayesnet}
\usepackage{subcaption}
\usepackage{graphicx}
\usepackage{ulem}

\usepackage{hyperref}
\usepackage{tikz}
\usetikzlibrary{positioning, shapes.geometric}
\usepackage{pgfplots}

\title{State space partitioning based on constrained spectral clustering for block particle filtering}

\usepackage{authblk}
\author[1,2,3]{Rui Min}
\author[2,3]{Christelle Garnier}
\author[4]{Fran\c{c}ois Septier}
\author[1]{John Klein}
\affil[1]{Univ. Lille, CNRS, Centrale Lille, UMR 9189 - CRIStAL, F-59000 Lille}
\affil[2]{IMT Nord Europe, Institut Mines-T\'el\'ecom, Centre for Digital Systems, F-59000 Lille}
\affil[3]{Univ. Lille, CNRS, Centrale Lille, Institut Mines-T\'el\'ecom, UMR 9189 - CRIStAL, F-59000 Lille}
\affil[4]{Universit\'e Bretagne Sud, LMBA, UMR CNRS 6205, F-56000 Vannes}
\date{}                     
\begin{document}

\maketitle

\begin{abstract}
The particle filter (PF) is a powerful inference tool widely used to estimate the filtering distribution in non-linear and/or non-Gaussian problems. To overcome the curse of dimensionality of PF, the block PF (BPF) inserts a blocking step to partition the state space into several subspaces or blocks of smaller dimension so that the correction and resampling steps can be performed independently on each subspace. Using blocks of small size reduces the variance of the filtering distribution estimate, but in turn the correlation between blocks is broken and a bias is introduced.

When the dependence relationships between state variables are unknown, it is not obvious to decide how to split the state space into blocks and a significant error overhead may arise from a poor choice of partitioning. 
In this paper, we formulate the partitioning problem in the BPF as a clustering problem and we propose a state space partitioning method based on spectral clustering (SC). We design a generalized BPF algorithm that contains two new steps: (i) estimation of the state vector correlation matrix from predicted particles, (ii) SC using this estimate as the similarity matrix to determine an appropriate partition. In addition, a constraint is imposed on the maximal cluster size to prevent SC from providing too large blocks.
We show that the proposed method can bring together in the same blocks the most correlated state variables while successfully escaping the curse of dimensionality. 
\end{abstract}

\textbf{keywords}:

\texttt{Block particle filter, high dimensional models, state space partitioning, constrained spectral clustering}


\section{Introduction}
\label{sec:Intro}

The filtering problem consists in estimating unknown quantities, referred to as hidden states, from observed data and by leveraging a state space model describing the dynamics of states and their links to observations. This problem arises in various domains such as time series analysis \cite{zhang2007particle}, statistical signal processing \cite{doucet2005monte}, automatic control systems \cite{rigatos2010extended} and geoscience applications \cite{wikle2007bayesian} where this problem is known as data assimilation. The particle filter (PF) \cite{gordon1993novel, doucet2001introduction} is a powerful inference tool widely used to estimate the filtering distribution in non-linear and/or non-Gaussian cases. It approximates the filtering distribution by a set of weighted samples called particles. However PF is poorly efficient in high dimensional state spaces because only a small fraction of samples drawn from the importance density have significant weights with respect to the observations \cite{rebeschini2015can}. To avoid weight degeneracy, PF requires to increase exponentially the number of particles with the dimension \cite{snyder2008obstacles, bickel2008sharp}, which is computationally too expensive.

Many methods have been proposed to overcome this limitation known as the curse of dimensionality, some reviews of these methods are available in \cite{septier2015overview,farchi2018comparison}. One promising class of methods is based on the localization principle. Localization exploits the low correlation observed between a majority of state variables and observations in high dimensional real systems, which can therefore be represented as sets of locally low dimensional models. By partitioning the state space into subspaces and introducing conditional independence assumptions, some filtering steps can be dispatched in the smaller dimensional subspaces. Such local methods have been developed in \cite{rebeschini2015can, djuric2013particle, min2021block, djuric2007multiple,closas2012improving, mihaylova2012parallelized}. 

Based on the localization principle, the block particle filter (BPF) proposed by Rebeschini and Van Handel \cite{rebeschini2015can} approximates the filtering distribution by the product of marginals on each subspace, called block. The framework of the bootstrap PF algorithm is kept. The prediction step is implemented in the usual way. Then a blocking step is inserted to partition the state space into blocks and the correction and resampling steps are performed independently on each block. 
Using blocks of small size reduces the variance of the filtering density estimate, but in turn the correlation between blocks (mainly at their boundaries) is broken and a bias is introduced. 
For a given block size, this bias does not vanish as the number of particles tends to infinity meaning that the estimator is not consistent. Therefore, as the number of particles increases, the block size should be increased in order to keep the asymptotic consistency of the BPF. In practice, for a finite number of particles, the appeal of BPF is that the variance reduction generally prevails over the possible bias increase. Consequently, the BPF performs much better than the standard PF in high dimensional problems in which localization assumptions are true.

However, the performance gain depends on the state space partition implemented in the BPF, including the size and composition of blocks. In \cite{rebeschini2015can}, the authors provide an appropriate order of magnitude for the block size given a number of particles, but they provide no method to decide how to split the state space into blocks. In a previous work \cite{min2021parallel}, we circumvented this issue by investigating a parallelization scheme that relies on multiple pre-selected state space partitions, each of which is used by one BPF instance. BPF instances are run in parallel and each one receives an equal fraction of the number of particles. On average, the parallel BPF outperforms a single BPF using all particles but only one randomly chosen partition. However it might be outperformed by a single BPF using one cleverly selected partition. 

The present paper investigates this line of thought and directly tackles the question of splitting the state space. We propose a partitioning method using constrained spectral clustering (CSC) to provide an appropriate partition to use in the BPF. To mitigate the correlation losses due to the blocking step, the aim is to bring together in the same blocks the most correlated components of the state vector. Thus, the state space partitioning problem in the BPF is revisited as a clustering problem, in which clusters are the desired blocks, state variables are seen as the data points to be clustered and correlation among state variables is used to quantify the similarity between data points. We design a generalized BPF algorithm that contains two new steps between the prediction and correction stages: (i) estimation of the state vector correlation matrix from predicted particles, (ii) spectral clustering (SC) to determine which state variables must be grouped together in the same blocks. However, clustering the most correlated variables in the same block is not enough because optimal partitions with respect to correlation might contain large clusters on which BPF would be less efficient. To avoid the resurgence of the curse of dimensionality, it is necessary to introduce a constraint on the maximal cluster size in SC.




The paper is organized as follows. In Section \ref{sec:BPF}, the filtering problem and the BPF are described. Section \ref{sec:SC} explains the proposed state space partitioning method based on constrained spectral clustering. Numerical results are shown in Section \ref{sec:simulation} and conclusions are given in Section \ref{sec:conclusion} .

\section{Bayesian filtering and the block particle filter}
\label{sec:BPF}

In this section, we define the filtering problem and its underlying representation by a state space model. We also give a reminder on the estimation of the filtering distribution by particle filtering (PF). Finally, we present the block particle filter (BPF) proposed by Rebeschini and Van Handel \cite{rebeschini2015can} to alleviate the PF curse of dimensionality. 

\subsection{The filtering problem}


The filtering problem consists in estimating an unknown and non-observable current state through a sequence of current and past state-dependent observations and a state space model describing state transition as well as state/observation relationship. This model $(\textbf{x}_{t}, \textbf{y}_{t})_{t \in \mathbb{N}}$, also called Hidden Markov Model (HMM) \cite{jazwinski2007stochastic}, is given by:
\begin{align}
\begin{split}
\mathbf{x}_{t} = f_{t}\left(\mathbf{x}_{t-1}, \mathbf{w}_{t}\right),
    \label{eqa:state}
\end{split}\\
\begin{split}
\mathbf{y}_{t} = h_{t}\left(\mathbf{x}_{t}, \mathbf{v}_{t}\right),
    \label{eqa:observation}
\end{split}
\end{align}
where equations (\ref{eqa:state}) and (\ref{eqa:observation}) are respectively the transition or state equation and the observation or measurement equation. $\mathbf{x}_{t}\in {} \mathbb{R}^{d_x}$ is the hidden state vector, $\mathbf{y}_{t}\in {} \mathbb{R}^{d_y}$ is the observation vector, $\mathbf{w}_{t}$  and $\mathbf{v}_{t}$ are respectively the state and observation noises, $f_{t}$ and $h_{t}$ are functions which can be non linear. The components of $\mathbf{x}_{t}$ are called state variables and denoted by ${x}_{t}(n )$ with $n \in  \lbrace 1:d_x \rbrace$. The components of $\mathbf{y}_{t}$ are called observations and denoted by ${y}_{t}(n)$ with $n \in  \lbrace 1:d_y \rbrace$.

Under quadratic loss, the minimizer of the expected posterior loss is the conditional expectation of $\mathbf{x}_{t}$ given all observations $\mathbf{y}_{1:t}$. It is thus necessary to infer the posterior probability density function (pdf) denoted by $p\left(\mathbf{x}_{t}|\mathbf{y}_{1:t}\right)$ and also called the filtering distribution. The structure of the HMM offers conditional independence properties: (i) the current observation $\mathbf{y}_{t}$ is independent from previous observations $\mathbf{y}_{1:t-1}$ and states $\mathbf{x}_{0:t-1}$ given the current state $\mathbf{x}_{t}$ and (ii) the current state $\mathbf{x}_{t}$ is independent from observations $\mathbf{y}_{1:t-1}$ and states $\mathbf{x}_{0:t-2}$ given the previous state $\mathbf{x}_{t-1}$. Consequently the joint posterior pdf can be factorized as follows:
\begin{equation}
p\left(\mathbf{x}_{0:t}|\mathbf{y}_{1:t}\right) \propto p\left(\mathbf{x}_{0}\right) \prod_{n=1}^t p\left(\mathbf{x}_{n}|\mathbf{x}_{n-1}\right)p\left(\mathbf{y}_{n}|\mathbf{x}_{n}\right).
\label{postHMM}
\end{equation}

In the Bayesian approach, the filtering distribution 
$p\left(\mathbf{x}_{t}|\mathbf{y}_{1:t}\right)$ can be obtained iteratively in two steps. First the prediction step gives the distribution of the state at time $t$ given only the past observations:
\begin{equation}
    p\left(\mathbf{x}_{t} | \mathbf{y}_{1:t-1}\right) = \int p\left(\mathbf{x}_{t}| \mathbf{x}_{t-1}\right) p\left(\mathbf{x}_{t-1}|\mathbf{y}_{1:t-1}\right)
    \mathrm{d} \mathbf{x}_{t-1}.
    \label{eqa:prediction}
\end{equation}
Then the correction step updates the prediction by taking into account the current observation:
\begin{equation}   p\left(\mathbf{x}_{t}|\mathbf{y}_{1:t}\right) = \frac{p\left(\mathbf{y}_{t} | \mathbf{x}_{t}\right) p\left(\mathbf{x}_{t} | \mathbf{y}_{1:t-1}  \right)}{p\left(\mathbf{y}_{t}| \mathbf{y}_{1:t-1}\right)}.
    \label{eqa:update}
\end{equation}

Except in a few cases including linear Gaussian models, it is impossible to calculate analytically the filtering distribution. 

\subsection{The particle filter (PF)}

The particle filter (PF) is currently widely used to solve the filtering problem in non-linear and/or non-Gaussian cases. It is a sequential Monte Carlo simulation tool based on importance sampling that approximates target probability distributions with discrete random measures \cite{djuric2003particle}. 

The PF approximates the filtering distribution with an empirical measure given by a set of weighted samples called particles: 

\begin{equation}
    p\left(\mathbf{x}_{t} \mid \mathbf{y}_{1: t}\right) \approx \sum_{i=1}^{N_{p}} w_{t}^{(i)}  \delta_{\mathbf{x}_{t}^{(i)}} (\mathbf{x}_{t}),
\label{eqa:posterior}
\end{equation}

where $N_p$ is the number of particles, $\mathbf{x}_{t}^{(i)}$ with $i \in  \lbrace 1:N_p \rbrace$ are the samples (or particles),  $w_{t}^{(i)}$ with $i \in  \lbrace 1:N_p \rbrace$ are the importance weights and $\delta_{\mathbf{x}}(\cdot)$ is a Dirac delta mass located at $\mathbf{x}$.

Using importance sampling, particles $\mathbf{x}_{t}^{(i)}$ are drawn from a distribution $q\left(\mathbf{x}_{t} \mid \mathbf{x}_{0: t-1}, \mathbf{y}_{1: t}\right)$, referred to as the importance density. The consistency of the estimator is asymptotically preserved by  recursively computing importance weights as follows:






\begin{equation}
    w_{t}^{(i)} \propto  w_{t-1}^{(i)}  \frac{p\left(\mathbf{y}_{t} \mid \mathbf{x}_{t}^{(i)}\right) p\left(\mathbf{x}_{t}^{(i)} \mid \mathbf{x}_{t-1}^{(i)}\right)}{q\left(\mathbf{x}_{t}^{(i)} \mid \mathbf{x}_{0: t-1}^{(i)}, \mathbf{y}_{1: t}\right)}.\label{eq:weight_norm}
\end{equation}
The normalization constraint on weights consists in enforcing that $\sum\limits_{i=1}^{N_{p}}  w_{t}^{(i)} = 1$. 

Choosing $q\left(\mathbf{x}_{t} \mid \mathbf{x}_{0: t-1}^{(i)}, \mathbf{y}_{1: t}\right) := p\left(\mathbf{x}_{t} \mid \mathbf{x}_{t-1}^{(i)}\right)$, i.e. choosing the transition density as importance density allows to simplify the weight update \eqref{eq:weight_norm}, which boils down to $w_{t}^{(i)} \propto  w_{t-1}^{(i)} p\left(\mathbf{y}_{t} \mid \mathbf{x}_{t}^{(i)}\right)$. 

Regardless of the choice of importance density, after a certain number of iterations, only a few particles have significant weights, all the other normalized weights are very close to zero. It means that a large computational effort is devoted to updating particles, most of them being useless. This is known as the weight degeneracy problem. To mitigate this issue, several resampling methods have been proposed \cite{gordon1993novel, kitagawa1996monte, beadle1997fast, liu1998sequential}. They consist in eliminating small-weight particles and replicating instances of large-weight particles. In a PF, the resampling step can be performed systematically at every time step or only when the weight variance exceeds a given threshold. A PF using the transition density as importance density and a systematic resampling is referred to as the bootstrap PF.

In spite of these workarounds, PF becomes unfortunately inefficient in high dimensional spaces due to the curse of dimensionality. 
In high dimension, densities are highly concentrated on small portions of the state space \cite{djuric2013particle}. This inevitably makes sampling poorly efficient as any $\mathbf{x}_t^{(i)} \sim q \left(\mathbf{x}_t| \mathbf{x}^{(i)}_{0:t-1},\mathbf{y}_{1:t} \right)$ is most likely doomed to exhibit small density values in the numerator of \eqref{eq:weight_norm}.
To deal with this limitation, PF requires to increase exponentially the number of particles with the dimension \cite{snyder2008obstacles, bickel2008sharp}, which is computationally too expensive.

\subsection{The block particle filter (BPF)}

In high dimensional models, it is common that some state variables have very limited dependence w.r.t. some other observations. 
This is notably the case in many statistical mechanical systems. For example, imagine that the components of vector $\mathbf{x}_t$ correspond to the curvature of an elastic membrane at different locations on this membrane. If a mass warps the membrane the curvature of nearby locations will be highly correlated while they will be quasi-independent for locations far from each other. Such a phenomenon is often referred to as correlation decay (in space).

Rebeschini and Van Handel \cite{rebeschini2015can} exploit this local dependence pattern to design a local particle filter, known as the block particle filter (BPF). A blocking step is inserted between the prediction and correction steps of the usual bootstrap PF. This modification takes advantage of the decay of correlation which acts as a dissipation mechanism for approximation errors in space.




\subsubsection{BPF algorithm}

In BPF, the first step is the same prediction step as in the bootstrap PF, i.e. particles $\textbf{x}_{t}^{(i)}$ are sampled from $p\left(\mathbf{x}_{t} \mid \mathbf{x}_{t-1}^{(i)}\right)$. Then a blocking step is inserted. 
The core idea of BPF relies on a partitioning of the state space into a set of $K$ independent and non overlapping blocks on which the correction and resampling steps are performed independently. 

Let us formally define the notion of block. If $K\leq d_x$, the state vector $\textbf{x}_{t}$ can be seen as the concatenation of $K$ subvectors followed by a permutation on entries:
\begin{equation}
\textbf{x}_{t}^{\top} = \sigma \left([\textbf{x}_{t,1}^{\top} \: \textbf{x}_{t,2}^{\top} \dots \textbf{x}_{t,K}^{\top}]\right)
\label{subvectors}
\end{equation}
where $\textbf{x}_{t,k}= \left[ x_{t}(n) : n \in B_{k}  \right]$ is the subvector containing state variables whose indices are in the subset $B_k \subseteq \lbrace 1:d_x \rbrace$
and $\sigma$ is a permutation of the set of all entries $x_{t}(n)$ which restores the ascending order of the indices $n\in \lbrace 1:d_x \rbrace$.
The subsets of indices $\lbrace B_{k} \rbrace_{k=1}^{K}$ are called \textbf{blocks} and verify: $\sqcup_{k=1}^{K} B_{k} =  \lbrace 1:d_x \rbrace$ where $\sqcup$ is the disjoint union (meaning $B_{k}\cap B_{k'} = \emptyset$, $\forall k\neq k'$). For simplicity, we will interchangeably also refer to $\textbf{x}_{t,k}$ as the $k^{\text{th}}$ block of $\textbf{x}_{t}$. The distinction between subvectors and subsets of indices will be made clear by the context. 
The partitioning may evolve over time, but here we omit the index $t$ to simplify notations. Such a partition into $K$ blocks is denoted by $\mathcal{P}_K=\{ B_1,  B_2, \cdots, B_K  \}$. The blocking step consists in performing this partition into subvectors.

An important locality assumption for the BPF to successfully circumvent the curse of dimensionality is that different subsets of observations are dependent on different state variables. More precisely, in this work, we assume the following factorization for the likelihood function:
\begin{equation}
p\left(\textbf{y}_{t} | \textbf{x}_{t}\right)= \prod_{n=1}^{d_x} \alpha_{t,n}\left(\textbf{y}_{t} , {x}_{t}(n)\right),\label{eq:likeli_factor2}
\end{equation}
for appropriate functions $\alpha_{t,n}(\cdot)$. For example, if $d_y=d_x$, this general representation can take into account models in which (i) each observation is only a function of one state variable: $\alpha_{t,n}\left(\textbf{y}_{t} , {x}_{t}(n)\right)=p({y}_{t}(n)|x_t(n))$ or (ii) there exists some correlation amongst observations: $\alpha_{t,n}\left(\textbf{y}_{t} , {x}_{t}(n)\right)=p({y}_{t}(n)|x_t(n),{y}_{t}(1),\cdots,{y}_{t}(n-1))$.



The factorization \eqref{eq:likeli_factor2} is exploited in the BPF to dispatch the correction step on each block. More precisely, after prediction and blocking, the subvector $\textbf{x}_{t,k}^{(i)}$ is updated by a local (block dependent) weight $w_{t,k}^{(i)}$ computed only from the observations related to the states of that block, i.e. $w_{t,k}^{(i)} \propto \prod_{n\in B_k} \alpha_{t,n}\left(\textbf{y}_{t} , {x}_{t,k}^{(i)}(n)\right)$. Finally, resampling is implemented per block and the locally resampled particles are re-assembled for the next prediction step.

Using the local weights, the BPF separately approximates $K$ marginal densities $\pi_{t,k}$ instead of the filtering distribution $p\left(\textbf{x}_{t}|\textbf{y}_{1:t}\right)$. 
At the end of each iteration, the BPF approximates $p\left(\textbf{x}_{t}|\textbf{y}_{1:t}\right)$ by the product of particle approximations of the marginals on each block:
 \begin{equation}
\hat{\pi}_t=\bigotimes_{k=1}^{K} \hat{\pi}_{t,k} \underset{\scriptscriptstyle N_p \rightarrow \infty }{\longrightarrow} \pi_t, 
\label{posterior_approx_1}
\end{equation} 
where the limiting target distribution $\pi_t$ is a biased approximation of $p\left(\textbf{x}_{t}|\textbf{y}_{1:t}\right)$. The bias vanishes to $0$ when the number of blocks $K$ tends to $1$ \cite{rebeschini2015can}. The estimation of the whole state vector from the product distribution is implemented by independently estimating each block of the state vector using the Monte Carlo approximation on that block.
The BPF algorithm is summarized in Algo. \ref{alg : Block particle filter}. 

\begin{algorithm}[H]
\caption{Block PF for a given state space partition $\mathcal{P}_K$ 
}
\label{alg : Block particle filter}
\begin{algorithmic}
\STATE \textbf{Inputs}: Partition into $K$ blocks: $\mathcal{P}_K=\{ B_1,  B_2, \cdots, B_K  \}$, number of particles $N_p$,  densities $p\left(\textbf{x}_{t} | \textbf{x}_{t-1}\right)$, $p\left(\textbf{y}_{t} | \textbf{x}_{t}\right)$ and $p\left(\textbf{x}_{0} \right)$.
\STATE \textbf{Initialization}: At $t=0$ sample particles $\textbf{x}_{0}^{(i)} \sim p\left(\textbf{x}_{0}\right)$, $\forall i \in [N_p]$ 
and set weights $w_{0, k}^{(i)}=\frac{1}{N_{p}}$, $\forall i \in [N_p]$ and $\forall k \in [K]$.
\STATE \textbf{Sequential Processing}: At $t\geq 1$ do
\STATE \textbf{1. Prediction step:}
\STATE Sample particles $\textbf{x}_{t}^{(i)} \sim p\left(\textbf{x}_{t} | \textbf{x}_{t-1}^{(i)}\right), \forall i \in [N_p]$ 
\STATE \textbf{2. Blocking/Correction step:}
\FOR{$k = 1: K$}
\STATE Get block weights  $w_{t, k}^{(i)} =\prod_{n\in B_k} \alpha_{t,n}\left(\textbf{y}_{t} , {x}_{t,k}^{(i)}(n)\right)$ , $\forall i \in [N_p] $ 
\STATE Normalize block weights
\ENDFOR
\STATE \textbf{3. Estimation step:}
\STATE Approximate the filtering distribution by: 
\\
\qquad $\hat{\pi}_t = \bigotimes_{k=1}^K \hat{\pi}_{t,k} = \bigotimes_{k=1}^K \sum_{i=1}^{N_{p}} w_{t, k}^{(i)} \delta_{\textbf{x}_{t, k}^{(i)}}(\textbf{x}_{t, k})$
\STATE \textbf{4. Resampling step:}
\FOR{$k = 1: K$}
\STATE Resample $\{\textbf{x}_{t, k}^{(i)}\}_{i=1}^{N_{p}} \sim \sum_{i=1}^{N_{p}} w_{t, k}^{(i)} \delta_{\textbf{x}_{t, k}^{(i)}}(\textbf{x}_{t, k})$
\ENDFOR
\end{algorithmic}
\end{algorithm}

\subsubsection{BPF performance}
\label{SectionBPFPerformance}

For a specific class of state space models, often referred to as locally interacting Markov models, Rebeschini and Van Handel provide an error bound for the BPF \cite{rebeschini2015can}. They consider a spatially extended model where $\textbf{x}_{t}$ and $\textbf{y}_{t}$ are vectors of the same dimension $d_x = d_y$ and whose entries correspond to space locations. The assumptions are as follows:
\begin{itemize}
\item $y_{t}(n)$ only depends on $x_{t}(n)$ and thus the likelihood function factorizes as: $p\left(\textbf{y}_{t} | \textbf{x}_{t}\right)= \prod_{n=1}^{d_x} p\left({y}_{t}(n) | {x}_{t}(n)\right)$.
\item $x_{t}(n)$ only depends on the entries of the previous state vector $\textbf{x}_{t-1}$ in the spatial neighborhood of $n$ with radius $r>0$ : 
$\mathcal{N}_{r}(n)= \lbrace m \in \lbrace 1:d_x \rbrace : |m-n| \leq r\rbrace$. Then the transition density factorizes as: 
$p\left(\textbf{x}_{t} | \textbf{x}_{t-1}\right)= \prod_{n=1}^{d_x} p\left({x}_{t}(n) | \textbf{x}_{t-1}\left(\mathcal{N}_{r}(n)\right)\right)$.
\end{itemize}

Given a partition $\mathcal{P}_K=\{ B_1,  B_2, \cdots, B_K  \}$, for any index (or locus) $n$ of a block $B_{k}$, the authors derive an error upper bound in the form of a bias and a variance: 
\begin{equation}
\| p\left(\textbf{x}_{t}|\textbf{y}_{1:t}\right)  - \hat{\pi}_t \|_n \leq \alpha \left[ \exp{\left(-\beta_1 \min_{m\in \partial B_{k}} |m-n|\right)} + \frac{\exp{\left(\beta_2 \:\max_k|B_{k}|\right)}}{\sqrt{N_p}}  \right]
\end{equation}


where the constants $\alpha$, $\beta_1$, $\beta_2 > 0$ depend on neighborhood size $r$ but depend neither on time $t$ nor on model dimension $d_x$, $\partial B_{k} = \lbrace n \in B_{k} : \mathcal{N}_{r}(n) \nsubseteq  B_{k} \rbrace$ is the subset of indices inside $B_{k}$ which can interact with components outside $B_{k}$ in one prediction step due to the state dynamics. 
To be precise, the norm $\| . \|_n$ is a local norm for some random measures $\rho$ (with law $\mathcal{L}$) defined as: 
\begin{equation}
    \| \rho \|_n = \underset{f\in \mathcal{F}^n, |f| \leq 1}{\sup} \sqrt{\mathbb{E}_{\rho \sim \mathcal{L}} \left[ \mathbb{E}_{\mathbf{x} \sim \rho} \left[ f(\mathbf{x}) \right] \right]},
\end{equation}
where $\mathcal{F}^n$ is the set of bounded measurable functions on $\mathbb{R}^{d_x}$ that coincide on every component of $\mathbf{x}$ except $x(n)$.

On the one hand, using blocks of small dimension (with few components) and/or a great number of particles can significantly reduce the variance of the filtering density estimate. On the other hand, using large blocks allows the limitation of the bias. This bias is introduced in the estimate by the blocking step because the correlation between blocks (mainly at the boundaries) is broken and the approximation (\ref{posterior_approx_1}) does not converge to the exact filtering distribution even as the number of particles goes to infinity. In practice, a trade-off is required. 

In cases where the notion of proximity between state variables corresponds to the shortest path length in a $q$-dimensional lattice,
the authors compute an upper bound for the average error over all indices and derive a rule of thumb to help tuning the block size for a given number of particles: $|B_k| \approx \log^{\frac{1}{q}}{N_p}, ~ \forall k \in \lbrace 1:K \rbrace$. In this paper, we will run experiments only on one-dimensional lattice settings ($q=1$). 


\subsubsection{Impact of the partitioning scheme}

\begin{figure}[H]
\centering
\begin{subfigure}[b]{0.49\textwidth}
         \centering
         \includegraphics[scale=.49]{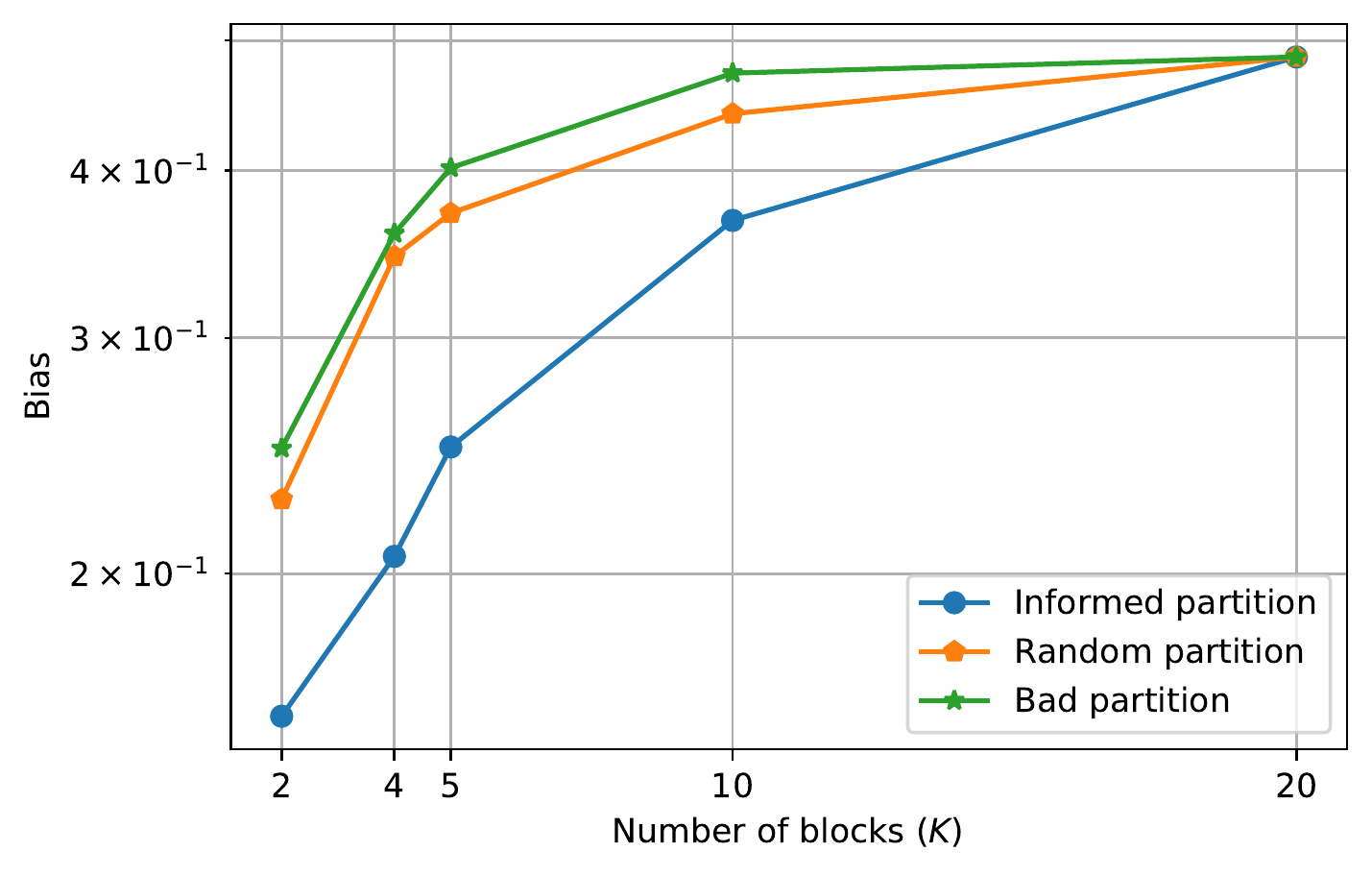}
    \label{fig:bias}
\end{subfigure}
\begin{subfigure}[b]{0.49\textwidth}
         \centering
    \includegraphics[scale=.49]{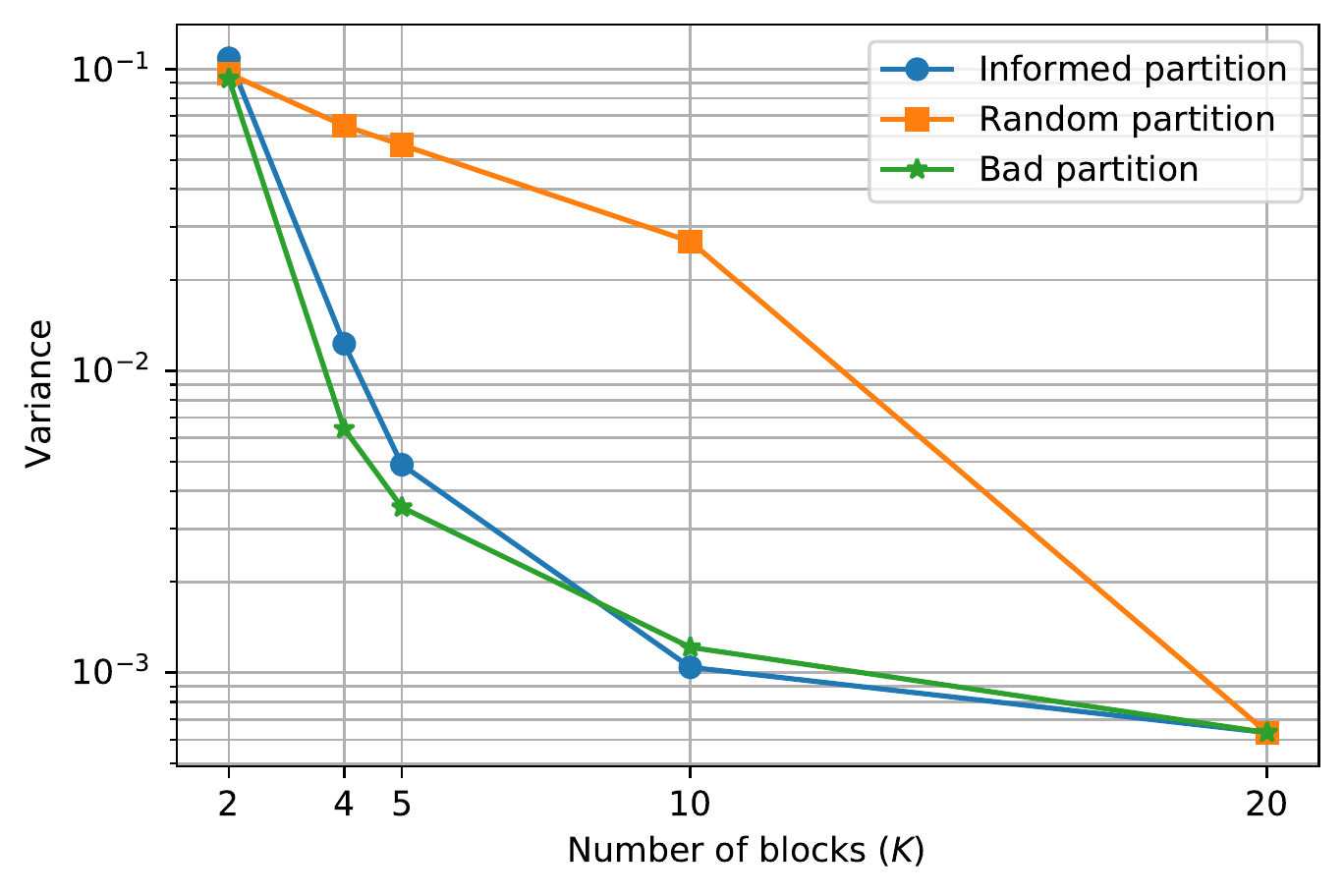}
    \label{fig:var}
\end{subfigure}
\caption{Bias (left) and variance (right) of BPFs versus the number of blocks $K$ (partitions into blocks of the same size). Case of a linear Gaussian model with a correlated state noise, $d_{x}= d_{y} = 20$, $N_p = 2000$ particles. }
\label{fig:partition}
\end{figure}

In practice, it appears that performance does not only depend on the block size but also on the composition of blocks, i.e. on the partition $\mathcal{P}_K$. Figure \ref{fig:partition} shows the bias and variance of the state estimates provided by BPFs using different partitions versus the number of blocks $K$ for a linear Gaussian model with a correlated state noise. This model is fully described in subsection \ref{ssec:linGaussModel}. In this experiment, the entries of the state noise covariance matrix $\mathbf{Q}$ are defined as: 
$$ Q\left(i,j\right) = \exp \left(-\frac{\left(i-j \right)^2}{100} \right) $$
The covariance between two state variables exponentially decreases when the gap between their indices increases. The model dimension is $d_{x}= d_{y} = 20$ and the state and observation vectors are split into blocks of the same size. Results are obtained with $N_p = 2000$ particles from $200$ simulations of length $50$ time steps. For comparison, three state space partitioning schemes are considered: 
\begin{itemize}
    \item a random partition scheme that randomly draws state variables at each time step to form blocks, 
    \item an ``informed partition" scheme that knows the structure of the state noise covariance matrix and thus chooses consecutive state variables for each block,
    \item a ``bad partition" scheme that assigns equally spaced components to each block. Intuitively it is a poor choice as blocks will be likely to contain state variables with very low pairwise levels of correlation thereby making the blocking step much more harmful.
\end{itemize}

As expected for a sufficient number of particles, the bias increases with the number of blocks, i.e. as the block size gets smaller, because the blocking step causes a greater loss of correlation, mainly at the boundaries. With the informed partition that gathers the most correlated components in the same blocks, the correlation loss is limited and the incurred bias is always smaller than those of other schemes. On the opposite, the bias is more severe when the BPF uses the bad partition that groups together very weakly correlated components. With regard to the variance, a decrease can be observed when using a greater number of blocks, i.e. blocks of smaller dimension. This is where the benefits of BPF can be seen in terms of contending with the curse of dimensionality. Note that the decrease is higher with the informed and bad partitions than with the random partition. We believe the randomness of this latter contributes to the overall observed state estimate variance. 

These simulation results highlight the impact of the partitioning scheme on the BPF performance. 
Rebeschini and Van Handel did not investigate ways to establish splitting strategies of the state space into blocks that could help reducing estimation error. The present paper is an attempt to fill this gap. The next section presents a methodology that leverages spectral clustering to identify relevant block partitions based on state variable correlations. 

\section{A new state space partitioning method for block particle filtering}
\label{sec:SC}


In this section, we propose a new state space partitioning method to automatically provide an appropriate partition to use in the BPF. To mitigate the correlation losses due to the blocking step, the goal is to divide the state vector into several blocks such that the most correlated variables are within the same block and the least correlated variables are in different blocks. Thus the state space partitioning problem in the BPF can be revisited as a clustering problem, in which state variables are seen as data points and clusters are blocks. The correlation among variables is used to quantify similarity between ``data points". 

To solve this problem, we propose to use a spectral clustering (SC) algorithm \cite{alpert1999spectral,malik2001contour}. Compared to traditional clustering algorithms, such as $K$-means or hierarchical clustering, SC has many advantages. SC performs an instrumental representation learning before actually clustering data points in the newly learnt space using one of the aforementioned traditional clustering methods. It can be thus seen as a generalization of these latter. SC learns a new representation in an unsupervised fashion by building upon a similarity measure. The definition of such a similarity measure is very flexible as it just have to be non-negative, symmetric and to achieve a maximal value when one point is compared with itself. This flexibility allows typically SC to find clusters that live in smaller dimensional non-linear manifolds. SC has thus the ability to detect clusters with complex shapes.

 In the following paragraphs, we introduce a generalized BPF algorithm that contains two new steps between the prediction and correction stages: estimation of the state vector correlation matrix from predicted particles and state space partitioning by SC using the estimated correlation matrix viewed as a state-component similarity matrix\footnote{Pairwise similarities among data points can be conveniently stored in a matrix.}. Furthermore, we implement a constrained SC (CSC) algorithm. Indeed, it is necessary to introduce a constraint on the maximal block size to avoid large blocks and accordingly the resurgence of the curse of dimensionality.

\subsection{Estimation of the state vector correlation matrix}

The first step of our state space partitioning method aims to estimate the state vector correlation matrix that will then be used as similarity matrix for SC. In the BPF, the set of particles (samples associated with weights) is available for the estimation. Each particle represents a possible realization of the state and each weight evaluates to what extent the particle is a good candidate state vector.  
A closer look at the BPF shows that correlation between each sample variables is introduced by the transition density during the prediction step, while the correction step performed independently on each block causes a correlation loss between variables assigned to different blocks. This is why the state correlation matrix estimation should be based on predicted particles. 
Indeed, suppose that one wishes to use the resampled particles instead. These particles are obtained after one loop of the BPF is completed. But if variables of indices $i$ and $j$ belong to different blocks, the blocking step implies that the correlation matrix obtained from resampled particles will have a zero entry at position $(i,j)$ and therefore SC is likely to find the same partition as the one used in the blocking step. In other words, we will be stucked with the same partition at any time step $t$.

Here, after resampling and prediction steps, the $N_p$ predicted particles have the same weight $\frac{1}{N_p}$ and provide an asymptotically unbiased estimator of the covariance matrix of the predictive posterior pdf $p(\mathbf{x}_t|\mathbf{y}_{1:t-1})$ based on the sample covariance matrix:
\begin{equation}
    \hat{\mathbf{\Sigma}}_{\mathbf{x}, t} = \frac{1}{N_p - 1}\sum_{i=1}^{N_p }\left(\mathbf{x}_{t}^{(i)}-\bar{\mathbf{x}}_{t}\right)\left(\mathbf{x}_{t}^{(i)}-\bar{\mathbf{x}}_{t}\right)^{\top},
    \label{eq: covariance}
\end{equation}
where:
\begin{equation}
    \bar{\mathbf{x}}_{t}=\frac{1}{N_p}\sum_{i=1}^{N_p} \mathbf{x}_{t}^{(i)}.
\end{equation}
Estimating a covariance matrix from a sample of random vectors has been extensively studied, especially in high dimension, and reviews of estimation techniques are available in \cite{cai2016estimating, fan2016overview}. The sample covariance matrix is still the most widely used estimator. It performs well when the sample size is much greater than the dimension, estimation errors happen when the sample size is not much larger than the dimension and the covariance matrix estimate is singular when the sample size is smaller.
In our problem, the size of the covariance matrix (and the associated correlation matrix) is $d_x\times d_x$ where the state dimension $d_x$ is large when considering high dimensional state spaces. Here we assume that the BPF uses a number of particles $N_p$ at least 3 to 10 times larger than the state dimension $d_x$ according to the model structure, so the sample size is high enough to obtain a correct estimate using the sample covariance matrix. This estimator has also the advantage of requiring no assumption on the matrix structure.

The correlation matrix estimate is then obtained by:
\begin{equation}
\hat{\mathbf{C}}_{\mathbf{x}, t} =   \mathbf{D_{1/a}}   \hat{\mathbf{\Sigma}}_{\mathbf{x}, t}  \mathbf{D_{1/a}}  
\label{eq: correlation}
\end{equation}
where $\mathbf{D_{1/a}}$ is the diagonal matrix of the inverse of the estimated standard deviations: 
$\mathbf{D_{1/a}} = \textrm{diag}\left( \frac{1}{a_1},\frac{1}{a_2},\dots , \frac{1}{a_{d_x}} \right )$ with $a_i = \sqrt{\hat{\Sigma}_{\mathbf{x}, t} (i,i)}$. 
Although other more challenging situations for covariance matrix estimation are out of the scope of this paper, these situations can be usually dealt with using shrinkage techniques. In particular, the estimator proposed in \cite{ledoit2004well} allows to automatically set the shrinkage parameter to an optimal value.

\subsection{Spectral clustering for state space partitioning}
\label{sec:CSP}

In a second step, the estimate of the correlation matrix of the state variables is used as similarity matrix for SC. More precisely, the similarity matrix $\textbf{S}\in \mathbb{R}^{d_x \times d_x}$ must be non-negative and symmetric, so it is equal to the absolute value of the correlation matrix estimate: $\textbf{S} = \left\lvert \hat{\mathbf{C}}_{\mathbf{x},t} \right\rvert $.


As mentioned before, SC is an unsupervised learning approach based on two main steps: (i) learning a new representation of the data points and (ii) applying a usual clustering approach (such as $K$-means) on the transformed data points (see \cite{von2007tutorial} for a detailed presentation of spectral clustering). The originality of SC relies on the first step 
which has actually a more general purpose than clustering. It is a graph partitioning method based on a new representation of data points. 

In our case, the items we wish to cluster are the entries of the random state vector, each of which is a scalar random variable. These entries are embedded in vectors that are simply obtained from the rows of the following design matrix:
\begin{align}
    \mathbf{X} = \begin{bmatrix} x_{t}^{(1)}(1) & \hdots & x_{t}^{(N_p)}(1) \\ 
    \vdots & \ddots & \vdots \\
    x_{t}^{(1)}(d_x) & \hdots & x_{t}^{(N_p)}(d_x) \\ 
    \end{bmatrix}.
\end{align}

Let $\mathbf{z}_{t}(i)$ denote the $N_p$-dimensional vector obtained from the $i^{\textrm{th}}$ row of $\mathbf{X}$, i.e. 
$\left.\mathbf{z}_{t}(i)\right.^{\top} =\begin{bmatrix} x_{t}^{(1)}(i) & \hdots & x_{t}^{(N_p)}(i) \end{bmatrix}$. 
Given the set of data points $\mathbf{z}_{t}(1), \mathbf{z}_{t}(2), \dots,$ $\mathbf{z}_{t}(d_x)$,
and the similarity $s_{i j} = \left\lvert  \hat{{C}}_{\mathbf{x}, t} (i,j) \right\rvert \geq 0$  between all pairs of data points 
$\mathbf{z}_{t}(i)$ and $\mathbf{z}_{t}(j)$,
we can construct an undirected weighted graph $G=(V, E)$. Each vertex $v_i$ in this graph represents a data point
$\mathbf{z}_{t}(i)$ and the edge between two vertices $v_i$ and $v_j$ is weighted by $\omega_{i j} \geq 0$ obtained from $s_{i j}$. 
In the usual graph terminology, the matrix $\boldsymbol\Omega\in \mathbb{R}^{d_x \times d_x}$ such that $\Omega(i,j) = \omega_{i j}$ is referred to as the adjacency matrix of the graph. There are several ways to define $\boldsymbol\Omega$ from the similarity matrix $\textbf{S}$ as discussed in the next subsection. 

To change the representation of the $d_x$ data points
$\mathbf{z}_{t}(i)$ into new points 
$\tilde{\mathbf{z}}_{t}(i)$, other matrices must be defined. The first one is called the degree matrix $\textbf{D}$ and is the diagonal matrix of the degrees of the vertices: $\mathbf{D} = \mathbf{diag}\left(\textrm{deg}_1,\textrm{deg}_2,\dots , \textrm{deg}_{d_x} \right )$ where 
$\textrm{deg}_i = \sum\limits_{j=1}^{d_x} \omega_{i j}$ is the degree of the vertex $v_i$. The second one has very interesting properties and is called the graph Laplacian matrix. 
Two different graph Laplacians can be defined from the adjacency and degree matrices. The unnormalized graph Laplacian is defined as: 
\begin{equation}
    \textbf{L}=\textbf{D}- \boldsymbol\Omega
\label{unnormL}
\end{equation}

From $\textbf{L}$, the normalized symmetric graph Laplacian is defined as \cite{chung1997spectral}:
\begin{equation}
    \textbf{L}_{\text {sym }}=\textbf{D}^{-1 / 2} \textbf{L} \textbf{D}^{-1 / 2} = \textbf{I} - \textbf{D}^{-1 / 2} \boldsymbol\Omega \textbf{D}^{-1 / 2}
\label{sym}
\end{equation}
where $\textbf{I}$ is the identity matrix.

In order to obtain $K$ clusters, SC selects the $K$ eigenvectors of $\textbf{L}_{\text {sym}}$ associated to the $K$ smallest eigenvalues (ignoring the trivial constant eigenvector). 
The $d_x$-dimensional eigenvectors $\mathbf{u}_1, \mathbf{u}_2, \dots, \mathbf{u}_K$ are then concatenated as columns in a matrix $\textbf{U} \in \mathbb{R}^{d_x \times K}$ and each line of this matrix corresponds to a new (smaller dimensional) data point 
$\tilde{\mathbf{z}}_{t}(i) \in \mathbb{R}^K$.
The advantage of the new representation is to make the task of the clustering algorithm used in step (ii) easier. Indeed, the eigenvectors will typically exhibit staircase patterns encoding cluster indicator functions.  Finally, a simple and computationally efficient algorithm such as $K$-means will have no difficulty to find $K$ clusters among the transformed data points.

\subsection{Introduction of constraints on block sizes}
\label{sec:constraints}

In our block particle filtering context, a general drawback of clustering is that, for a given number $K$ of clusters, it can lead to a wide variety of block sizes and the curse of dimensionality of Monte-Carlo methods can reappear inside blocks with large sizes. To prevent SC from creating large clusters, we set an upper bound $\zeta$ on the cluster sizes $|B_k|$ in the clustering algorithm used in step (ii) of SC.
To take this constraint into account, one can rely on constrained $K$-means methods that have been proposed in \cite{bradley2000constrained, ganganath2014data}. In this paper, we use the constrained $K$-means algorithm described in \cite{bradley2000constrained}. 

At each iteration, the unconstrained $K$-means algorithm includes two steps (i) Cluster assignment: each data point is assigned to the cluster whose center is closest to that point \footnote{Closest in the sense of Euclidean distance, the cluster center being the average of data points belonging to it.} and (ii) Cluster update: each cluster center is updated based on the new assignments made in the previous step. From randomly initialized cluster centers, the algorithm iterates until convergence (which is provably achieved after a finite number of iterations). 
In \cite{bradley2000constrained}, the authors revisit the problem by introducing selection variables equal to $0$ or $1$ that encode the cluster membership of each data point. 
With this new formulation of $K$-means clustering, it becomes possible to introduce constraints on cluster sizes. In \cite{bradley2000constrained}, a lower bound $\xi_k$ is imposed on the size of each cluster (denoted by $B_k$ in our problem). 

More precisely, the cluster assignment step is rewritten as a Minimum Cost Flow (MCF) linear network optimization problem, for which dedicated algorithms can be readily used and return optimal selection variables in $\lbrace 0 , 1 \rbrace$. 
The MCF formulation is based on a graph structure with nodes and directed edges or arcs. Each data point is a source node with a supply of $1$, each cluster is a sink node with a negative supply (i.e. a demand) equal to $-\xi_k$, which means each cluster will receive at least $\xi_k$ data points. A directed arc is defined from each source node to each sink node. The associated cost is the squared Euclidian distance between the corresponding data point (in our case $\tilde{\mathbf{z}}_t(i)$) and the corresponding cluster center. To ensure that all sources will "ship" their supplies, an artificial final sink node is added with a supply equal to $\sum_{k=1}^{K} \left(-|B_k| + \xi_k \right) = -d_x + \sum_{k=1}^{K} \xi_k$ in our case ($d_x$ being the number of data points). Directed edges connecting each cluster node to the final node have a zero cost. 

Our partitioning method for BPF does not need to force minimal block sizes, so we set $\xi_k=\xi=1, \forall k$. 
However, it does need to impose a maximal block size. This constraint, denoted by $\zeta$, can be easily incorporated in the MCF formulation by defining a flow capacity (in term of upper bound) on the edges connecting cluster nodes to the final node. If a capacity of $\zeta-\xi$ is assigned to each edge, then a feasible solution of the MCF problem cannot "ship" more than $\zeta$ supplies to each cluster node. The constrained SC algorithm is summarized in Algo. \ref{alg : NSC}. 

\begin{algorithm}[H]
\caption{Constrained spectral clustering (CSC) 
}
\label{alg : NSC}
\begin{algorithmic}
\STATE \textbf{Inputs}: Adjacency matrix $\mathbf{\boldsymbol\Omega}$, number of clusters $K$, max block size $\zeta$.
\STATE 1. Compute the normalized symmetric Laplacian $\mathbf{L_{sym}}$ using Eq. \eqref{sym}.
\STATE 2. Compute the $K$ eigenvectors $\mathbf{u}_1, \mathbf{u}_2, \dots, \mathbf{u}_K$ of $\mathbf{L_{sym}}$ associated to the $K$ smallest eigenvalues. 
\STATE 3. Concatenate these vectors as columns to create the matrix $\mathbf{U} \in \mathbb{R}^{d_x \times K}$ and normalize the rows of $\mathbf{U}$ to unit norm. 
\STATE 4. Get the new points $\tilde{\mathbf{z}}_t(i) \in \mathbb{R}^{K}$ from the $i^{\textrm{th}}$ rows of $\mathbf{U}$ , $\forall i \in [d_x]$.
\STATE 5. Partition the new points $\{\tilde{\mathbf{z}}_t(i)\}_{i=1}^{d_x}$ into $K$ clusters $B_1,...,B_K$ using the $K$-means algorithm under the constraint $|B_k|\leq \zeta$ , $\forall k \in [K]$.
\end{algorithmic}
\end{algorithm}


\subsection{Block particle filter algorithm with adaptive partitioning}
\label{subsec:main_algo}

In the previous subsections, we have introduced all necessary ingredients to identify a meaningful state space partition allowing to use the BPF when the dependence structure between components of the state and observation vectors is not known. This new approach is summarized in Algo. \ref{alg : SP Block particle filter}.





Observe that in this algorithm, we use the most simple SC weight definition, i.e. $\omega_{ij} = s_{ij}, \forall i,j$. Indeed, other choices such as those involving a radial basis function kernel have proved to provide no performance improvement in our experiments. In addition, using such a kernel requires to tune a parameter controlling its width. Our choice thus avoids the burden of tuning this parameter. 

Besides, compared to the usual BPF, Algo. \ref{alg : SP Block particle filter} requires to tune correctly the maximal block size parameter $\zeta$. In the next section which contains numerical validation of this algorithm, one can check that our approach exhibits moderate sensitivity w.r.t. $\zeta$. As guiding rule for tuning this parameter, we recommend to use $\left\lceil \gamma \frac{d_x}{K}\right\rceil$ where $\gamma$ is a scalar in $[1,2]$ and $\left\lceil .\right\rceil$ is the ceiling function. Note that when $K$ is a divisor of $d_x$ and $\gamma=1$, then Algo. \ref{alg : SP Block particle filter} forces all blocks to have the same size. Setting $\gamma$ in $]1,2]$ gives more flexibility in clustering while limiting the block size. 

\begin{algorithm}[H]
\caption{Block PF with adaptive state space partitioning
}
\label{alg : SP Block particle filter}
\begin{algorithmic}
\STATE \textbf{Inputs}: Number of particles $N_p$, number of clusters/blocks $K$, max block size $\zeta$, densities $p\left(\textbf{x}_{t} | \textbf{x}_{t-1}\right)$, $p\left(\textbf{y}_{t} | \textbf{x}_{t}\right)$ and $p\left(\textbf{x}_{0} \right)$.
\STATE \textbf{Initialization}: At $t = 0$ sample particles $\textbf{x}_{0}^{(i)} \sim p\left(\textbf{x}_{0}\right)$ 
and set weights $w_{0, k}^{(i)}=\frac{1}{N_{p}},\forall i \in [N_p]$ and $\forall k \in [K]$.
\STATE \textbf{Sequential Processing}: At $t\geq 1$ do
\STATE \textbf{1. Prediction step:}
\STATE Sample particles $\textbf{x}_{t}^{(i)} \sim p\left(\textbf{x}_{t} | \textbf{x}_{t-1}^{(i)}\right), \forall i \in [N_p]$ 
\STATE \textbf{2. Partitioning step:}
\STATE Estimate the correlation matrix $\hat{\mathbf{C}}_{\mathbf{x}, t}$ using Eqs. \eqref{eq: covariance} and \eqref{eq: correlation}
\STATE Run CSC Algo. \ref{alg : NSC} with $\boldsymbol\Omega = \left\lvert \hat{\mathbf{C}}_{\mathbf{x},t} \right\rvert$ , $K$ , $\zeta$ as inputs to obtain the state space partition at time $t$:  $\mathcal{P}_{t,K}=\{ B_{t,1},  B_{t,2}, \cdots, B_{t,K}  \}$

\STATE \textbf{3. Blocking/Correction step:}
\FOR{$k = 1: K$}
\STATE Get block weights $w_{t, k}^{(i)} =\prod_{n\in B_{t,k}} \alpha_{t,n}\left(\textbf{y}_{t} , {x}_{t,k}^{(i)}(n)\right)$ , $\forall i \in [N_p] $ 
\STATE Normalize block weights
\ENDFOR
\STATE \textbf{4. Estimation step:}
\STATE Approximate the filtering distribution by: 
\\
\qquad $\hat{\pi}_t = \bigotimes_{k=1}^K \sum_{i=1}^{N_{p}} w_{t, k}^{(i)} \delta_{\textbf{x}_{t, k}^{(i)}} (\textbf{x}_{t, k})$
\STATE \textbf{5. Resampling step:}
\STATE Resample $\{\textbf{x}_{t, k}^{(i)}\}_{i=1}^{N_{p}} \sim \sum_{i=1}^{N_{p}} w_{t, k}^{(i)} \delta_{\textbf{x}_{t, k}^{(i)}} (\textbf{x}_{t, k})$ , $\forall k \in [K]$
\end{algorithmic}
\end{algorithm}
Another comment on this algorithm is that partitioning is adaptive when it is carried out at each time step. In some cases, practitioners may find it sufficient to compute the partition once at $t=1$ and stay with this partition for the rest of the BPF run. Of course this greatly reduces the computation time overhead incurred by step 2. of Algo. \ref{alg : SP Block particle filter}. 

Speaking of this overhead, we can estimate the complexity due to the additional partitioning step. The correlation matrix estimation has complexity $O\left(N_p d_x^2\right)$ which in many cases is less than the complexity of the prediction step. For spectral clustering (Algo. \ref{alg : NSC}), aside from step 5. which is a call to constrained $K$-means, the costliest operation is to find the eigenvectors. There is a variety of numerical methods that solve eigenvalue decomposition iteratively. To fix ideas, the time complexity of these methods should not exceed $O \left( d_x^3\right)$. 
Concerning the constrained $K$-means, the MCF formulation allows to find clusters in polynomial time as well. This time depends on:
\begin{itemize}
    \item the number of nodes in the MCF graph: $d_x + K + 1$,
    \item the number of arcs: $K\times d_x + K = \left(K+1\right)d_x$,
    \item the maximal cost, in absolute value, associated to an arc. Any arc from a source node to a cluster node has a cost equal to the squared Euclidean distance between the corresponding vector $\tilde{\mathbf{z}}_t(i)$ and the corresponding cluster center. Other arcs have zero cost. So the cost is upper bounded by $C= \underset{1 \leq i,j\leq d_x}{\max}\: \lVert \tilde{\mathbf{z}}_t{(i)} - \tilde{\mathbf{z}}_t(j) \rVert
   ^{2}_{2}$ where $\lVert . \rVert_{2}$ is the Euclidean norm in $\mathbb{R}^{K}$.
\end{itemize}
Given these numbers, the expected time for solving the MCF problem has complexity $O\left( \left(d_x + K + 1\right)^2\left(K+1\right)d_x \log\left(\left(d_x + K + 1\right)C\right) \right)$.


\section{Simulation results}
\label{sec:simulation}

This section provides experimental validation of the state space partitioning method that we propose to automatically provide an appropriate partition into blocks to insert in the BPF.
The experiments are carried out in high dimensional state spaces and highlight the benefits of state space partitioning using spectral clustering with a constraint on the maximal block size. 
We start with some simulations involving a linear Gaussian model. Next, we investigate a non-linear model: the Lorenz 96 model which induces chaotic evolution and correlation of the state variables. In both models, correlation is also introduced in the posterior distribution via a correlated state noise process. We compare different versions of the BPF:
\begin{itemize}
\item BPF with random partition, which uses a randomly selected partition at each time step, 
\item BPF with known partition, which uses the natural candidate partition, directly derived from the dependence pattern of the state process (assumed to be known),
\item BPF with unknown partition, which uses the partition provided by our partitioning method based on spectral clustering (the dependence pattern is a priori unknown). We consider unconstrained SC ($\zeta=d_x$) and constrained SC with a smaller or larger maximum block size under the guideline stated in subsection \ref{subsec:main_algo}: $\zeta=\left\lceil \gamma \frac{d_x}{K}\right\rceil$ with $\gamma \in [1,2]$.  
\end{itemize}

\subsection{Linear Gaussian model}
\label{ssec:linGaussModel}
Although the Bayes optimal solution in the linear Gaussian case is given by the Kalman filter (KF) \cite{kalman1960new}, this setting provides a very interesting playground to benchmark more general purpose filters such as particle filters. In this paper, the following linear Gaussian model is considered: 
\begin{align}
    \begin{split}
    \mathbf{x}_{t}& = \mathbf{F} \mathbf{x}_{t-1} + \mathbf{w}_{t}\\
     \mathbf{y}_{t} &= \mathbf{H} \mathbf{x}_{t} + \mathbf{v}_{t}
    \end{split}
     \label{linear}
\end{align}
where $\mathbf{F}$ is the transition matrix, $\mathbf{H}$ is the observation matrix, $\mathbf{w}_{t}$ and $\mathbf{v}_{t}$ are mutually independent Gaussian noises:  $ \mathbf{w}_{t} \sim \mathcal{N}(\mathbf{0}, \mathbf{Q})$, and $ \mathbf{v}_{t} \sim \mathcal{N}(\mathbf{0}, \mathbf{R})$. 
The initial state is $\textbf{x}_{0} \sim \mathcal{N}(\mathbf{0}, \mathbf{\Sigma}_{0})$. In this model, state and observation dimensions are $d_{x}= d_{y} = 100$, matrices $\mathbf{F}$, $\mathbf{H}$, $\mathbf{R}$ and $\mathbf{\Sigma}_{0}$ are equal to the identity matrix, and correlation is introduced by the state noise. In the following subsections, we consider different structures for the covariance matrix $\mathbf{Q}$ of the state noise. In any case, performance results are obtained from $100$ simulations of length $50$ time steps.

\subsubsection{State noise with a block diagonal covariance matrix}

First the state noise covariance matrix is assumed to be block diagonal. 
In this case, matrix $\mathbf{Q}$ (through its block diagonal structure) gives access to an obvious candidate partition $\mathcal{P}_K$ in which each block $B_k$ contains the set of indices corresponding to one block of the matrix. If blocks have small sizes, we can expect that this known or informed partition is a very good choice for BPF.

In order to compare this known partition and the partition issued from constrained spectral clustering, we use the Adjusted Rand Index (ARI) \cite{hubert1985comparing, steinley2004properties}, which is a similarity measure derived from the numbers of pairs of elements that are assigned in the same and different clusters in the two partitions. The value of ARI is in the range of $0$ to $1$, a value of $1$ corresponding to a perfect match between partitions. 

\paragraph{Covariance matrix blocks with identical small size} 

\begin{figure}[H]
     \centering
     \begin{subfigure}[b]{0.49\textwidth}
         \centering
         \includegraphics[height=4cm]{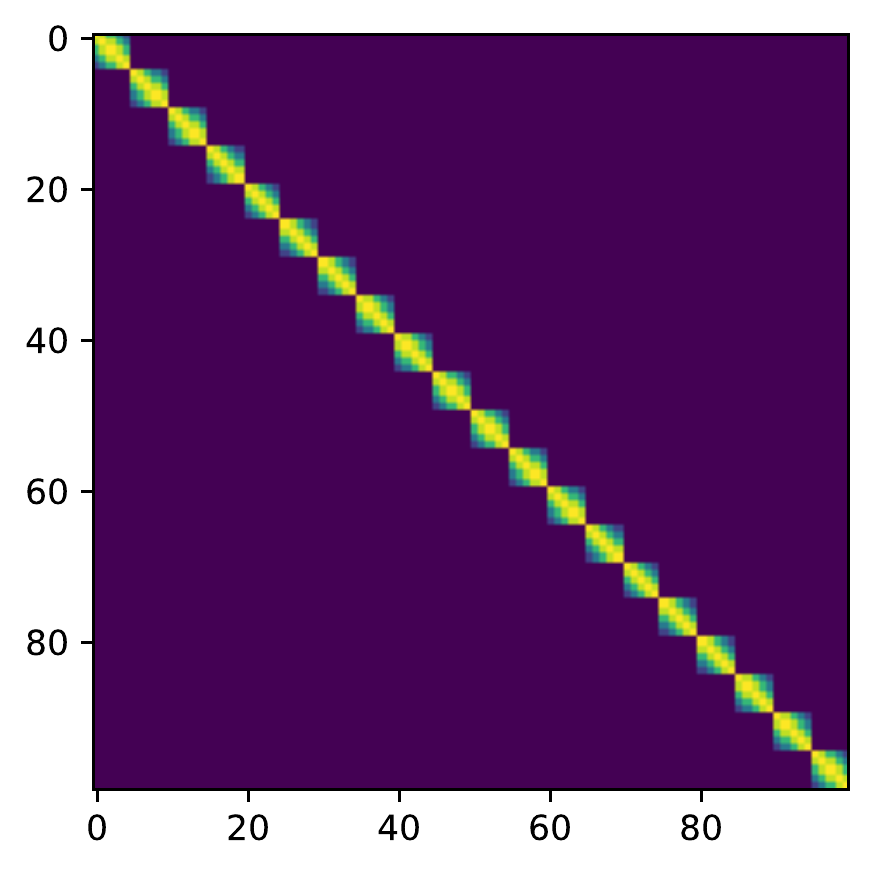}
         \caption{$l = 10$}
         \label{fig: l_10}
     \end{subfigure}
     \hfill
     \begin{subfigure}[b]{0.49\textwidth}
         \centering
         \includegraphics[height=4cm]{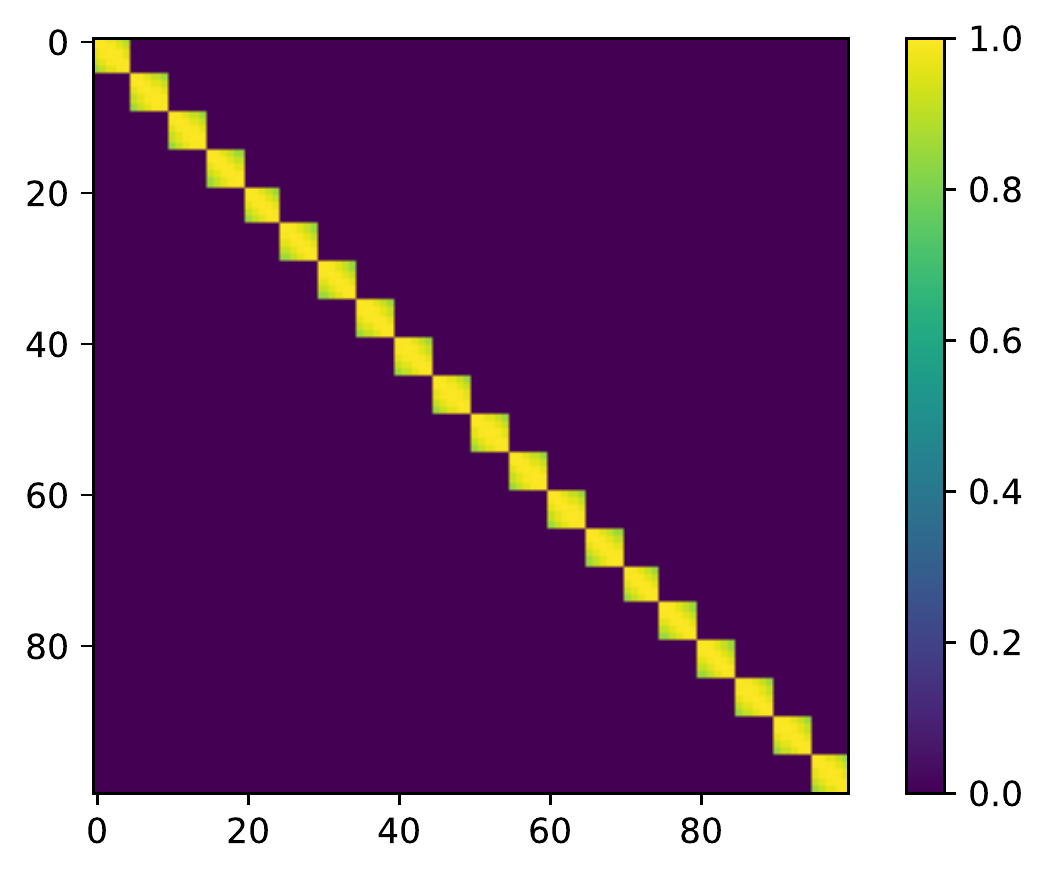}
         \caption{$l = 100$}
         \label{fig: l_100}
     \end{subfigure}
        \caption{Visualisation of the matrix $\mathbf{Q}$ for two values of $l$}
        \label{fig: Q_small}
\end{figure}

First we consider a state noise covariance matrix $\mathbf{Q}$ with small blocks of the same size. 
More precisely, $\mathbf{Q}$ has $20$ square matrices (blocks) of size $5$ in the main diagonal. Inside each block, the entries are given by 
$Q(i,j) = \exp{(-(i-j)^2 / l)}$. 
We can obtain different levels of correlation by changing the value of the parameter $l$, smaller values of $l$ meaning less correlation among the state variables within one block (see Fig. \ref{fig: Q_small} for a visualisation). 

In this case, we can expect the "known" partition in line with the covariance matrix structure, consisting of $K = 20$ blocks of size $|B_k|=5$ including consecutive entries, is the best choice for BPF. In this first experiment, only $N_p=100$ particles are used. In our generalized BPF with adaptive partitioning, the number of samples available for the state correlation matrix estimation is equal to the state dimension $d_x$. Therefore we can expect an imprecise estimate, making the task of spectral clustering not trivial. For the clustering step, we fix $K=20$ blocks and consider several values of the constraint $\zeta$ on the maximum block size: $\zeta=d_x$ (unconstrained SC), $\zeta=\left\lceil  \frac{d_x}{K}\right\rceil=5$ (partition with blocks of the same size $|B_k|=5$) and $\zeta=\left\lceil 1.5 \frac{d_x}{K}\right\rceil=8$ (more flexibility in the choice of the partition while restricting the block size).




\begin{figure}[H]
     \centering
     \begin{subfigure}[b]{0.49\textwidth}
         \centering
         \includegraphics[width=\linewidth]{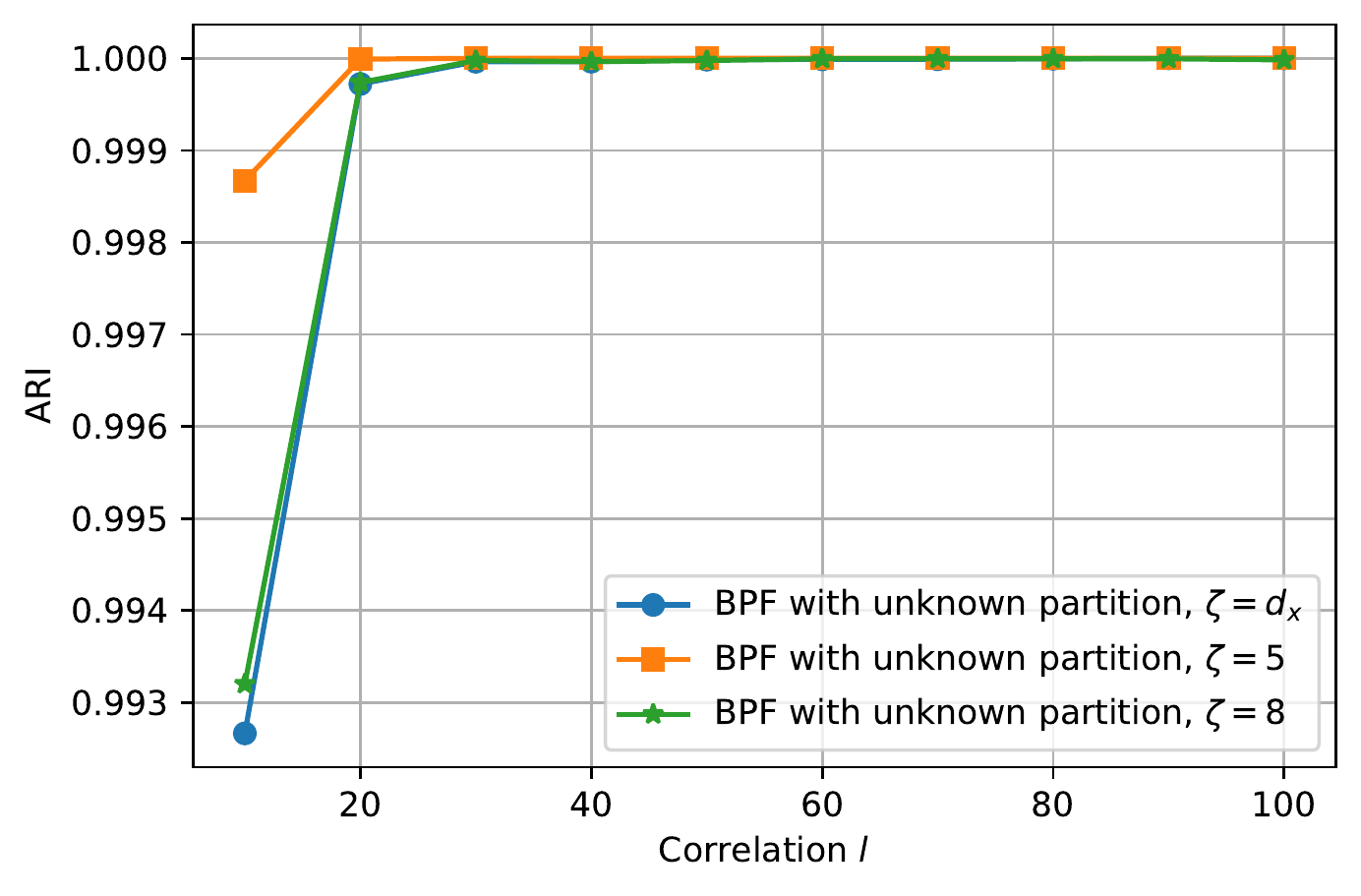}
         \caption{ARI}
         \label{fig: ARI_small}
     \end{subfigure}
     \hfill
     \begin{subfigure}[b]{0.49\textwidth}
         \centering
         \includegraphics[width=\linewidth]{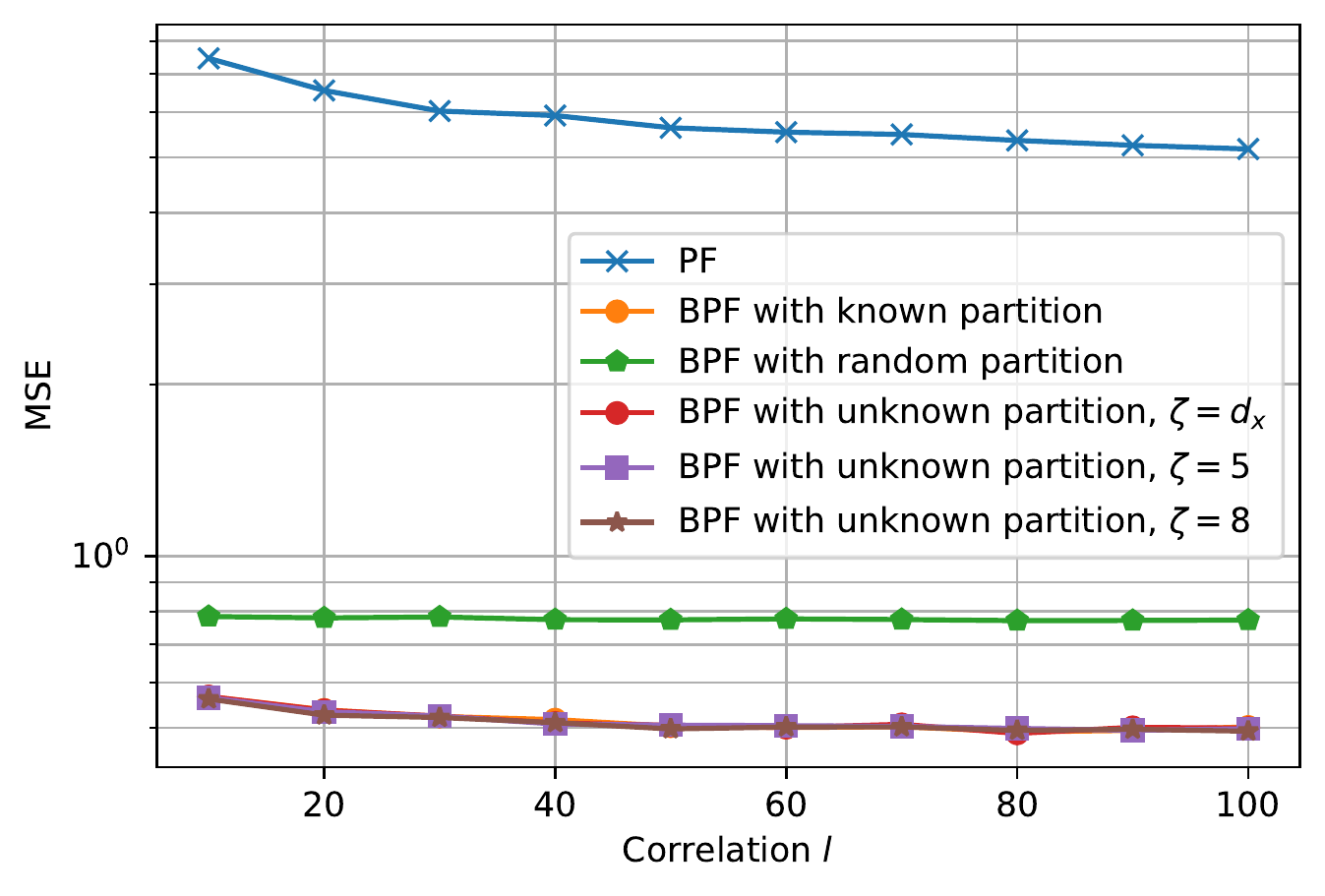}
         \caption{MSE}
         \label{fig: MSE_small}
     \end{subfigure}
        \caption{Performance of PF and BPFs (with $K=20$ blocks) versus the parameter $l$ in a linear Gaussian model for $N_p=100$. State noise with a block diagonal correlation matrix ($20$ blocks of the same size on the main diagonal).}
        \label{fig: small}
\end{figure}

Fig. \ref{fig: ARI_small} gives the ARI between the known partition, directly deduced from the structure of $\mathbf{Q}$, and some partitions obtained by the proposed partitioning scheme (step 2 of Alg. \ref{alg : SP Block particle filter}). Regardless of the value of the constraint $\zeta$, our method is able to find the exact partition in line with the covariance matrix pattern for $l\geq 30$. It also provides a very close partition for smaller values of $l$, clustering from low correlation being slightly more difficult. Despite the low number of particles, the state correlation matrix is estimated with sufficient precision for SC. 

Accordingly, Fig. \ref{fig: MSE_small} shows that there is no difference in performance in terms of mean squared error (MSE) between BPF exploiting the known partition and BPF learning this partition on the fly. It also shows that using this appropriate partition in BPF offers a performance gain compared to a random selection of the partition. 

\paragraph{Covariance matrix blocks with different time-varying sizes}

In this second experiment, we consider a slightly more complex model. The state noise covariance matrix $\mathbf{Q}$ has $10$ blocks with different sizes in the main diagonal. The block sizes are respectively equal to $5, 9, 8, 12, 13, 7, 15, 14, 11, 6$ for time steps $t = 1$ to $25$ and $8, 14, 11, 15, 12, 5, 13, 9 , 6, 7$ for $t = 26$ to $50$. 
Inside each block, the entries are given by 
$Q(i,j)= \exp{(-(i-j)^2 / l)}$ with $l=100$. Unlike the previous case, we can expect that the known partition directly derived from the structure of $\mathbf{Q}$ is not necessarily ideal because some blocks are rather large and the covariance matrix of the posterior distribution is not necessary block diagonal (especially just after $t=26$, the instant of the matrix change).
As previously, only $N_p=100$ particles are used. So in the BPF, the correlation matrix estimate obtained by step 2 of Algo. \ref{alg : SP Block particle filter} is likely to be noisy.


\begin{table}[h]
\centering
\small
\begin{tabular}{|c|c|c|}
\hline
 Bayesian filter    &  MSE & ARI\\
  \hline
  KF (optimal filter) & 0.2331 & /\\
  \hline
  Bootstrap PF & 4.2107 & / \\
  \hline
  BPF with known partition of 10 blocks & 0.8185 & / \\
  \hline 
  BPF with random partition (same size blocks) & 1.1466 & / \\
  \hline
  BPF with unknown partition ($\zeta=d_x$ / unconstrained SC)  & 0.8190 & 0.9938\\
  \hline
  BPF with unknown partition ($\zeta$ = 10 / same size blocks) & 0.7067 & 0.7010 \\
  \hline
  BPF with unknown partition ($\zeta$ = 12) & 0.7473 & 0.8701\\
  \hline
  BPF with unknown partition ($\zeta$ = 15) & 0.8070 & 0.9942\\
  \hline
  
\end{tabular}
\caption {Performance of KF, PF and BPFs (with $K=10$ blocks) in a linear Gaussian model for $N_p=100$. State noise with a time varying block diagonal correlation matrix ($10$ blocks with different sizes on the main diagonal, $l=100$).
\label{tab:lin_gauss1}}
\end{table}

\normalsize

MSE and ARI performances of different Bayesian filters are reported in Table \ref{tab:lin_gauss1}. The table shows that all versions of BPF perform a lot better than the bootstrap PF and among them, BPF with a random partition is less efficient. 

Regarding BPFs with partitions learnt on the fly, when the maximal size constraint is looser ($\zeta = 15$) or omitted ($\zeta = d_x$), ARI values are close to 1\footnote{Note that achieving an ARI of $1$ is not necessarily possible because, near $t=26$, the covariance matrix of the posterior predictive distribution is not perfectly block diagonal and SC may legitimately choose a partition different from the one deduced from $\mathbf{Q}$.}. Our partitioning method nearly retrieves the known partition in line with matrix $\mathbf{Q}$, whose blocks have a maximum size of $15$. Clustering from state correlation estimation is near-perfect, even with only $100$ particles. Accordingly, BPFs with both constraint values exhibit MSE close to the BPF with the known partition. 
When the constraint is stronger ($\zeta = 10$ or $12$), MSE decreases. Algo. \ref{alg : SP Block particle filter} partitions the state space into smaller blocks (ARI values deviate from $1$) and BPF is less prone to the curse of dimensionality. 
Smaller blocks necessarily means breaking some correlations and increasing the BPF bias. But in return, the variance term is reduced by a greater order of magnitude. In this case, the benefit of introducing a constraint in SC is confirmed. 



\begin{table}[h]
\centering
\small
\begin{tabular}{|c|c|c}
\hline
 Bayesian filter    &  MSE \\
  \hline
  KF (optimal filter) & 0.2331 \\
  \hline
  Bootstrap PF & 4.2107 \\
  \hline
  BPF with known partition of 10 blocks  & 0.8185 \\
  \hline 
  BPF with random partition (same size blocks) & 0.7573  \\
  \hline
  BPF with unknown partition ($\zeta=d_x$ / unconstrained SC)  & 0.4613\\
  \hline
  BPF with unknown partition ($\zeta$ = 5 / same size blocks) & 0.4745 \\
  \hline
  BPF with unknown partition ($\zeta$ = 8) & 0.4681 \\
  \hline
  
\end{tabular}
\caption {MSE performance of KF, PF and BPFs (with $K=20$ blocks) in a linear Gaussian model for $N_p=100$. State noise with a time varying block diagonal correlation matrix ($10$ blocks with different sizes on the main diagonal, $l=100$).\label{tab:lin_gauss2}}
\end{table}

\normalsize

If one wishes to follow the rule of thumb derived in \cite{rebeschini2015can} and presented in subsection \ref{SectionBPFPerformance}, block size $|B_k|$ should be around $\log N_p = \log 100 = 4.6$. If we choose $|B_k|=5$, then the recommended number of blocks is larger than $10$: $K=20$. 

Table \ref{tab:lin_gauss2} reports MSE performance of several BPFs for $K=20$. For comparison, the four first rows of Table \ref{tab:lin_gauss1} are also reported. ARI has no interest here because this setting does not allow SC to find the known partition originating from $\mathbf{Q}$. 
Obviously, partitioning into $K=20$ blocks and accordingly into smaller blocks induces more correlation losses and a higher BPF bias. But the variance reduction still prevails over the bias increase. Indeed, all versions of BPF, even the one using random partitions, perform much better than previously and outperform the BPF with the known partition into $10$ blocks deduced from $\mathbf{Q}$. BPFs using the proposed clustering technique achieve the best performance, regardless of the value of the constraint $\zeta$. Roughly speaking, MSE values are twice smaller compared to values reported in Table \ref{tab:lin_gauss1} for partitioning into $K=10$ blocks. For a larger value of $K$, i.e. smaller blocks, the constraint  $\zeta$ has coherently less impact on performance. 


\subsubsection{State noise with a dense covariance matrix}
\label{sssec:densecov}

\begin{figure}[H]
         \centering
         \includegraphics[width=5cm]{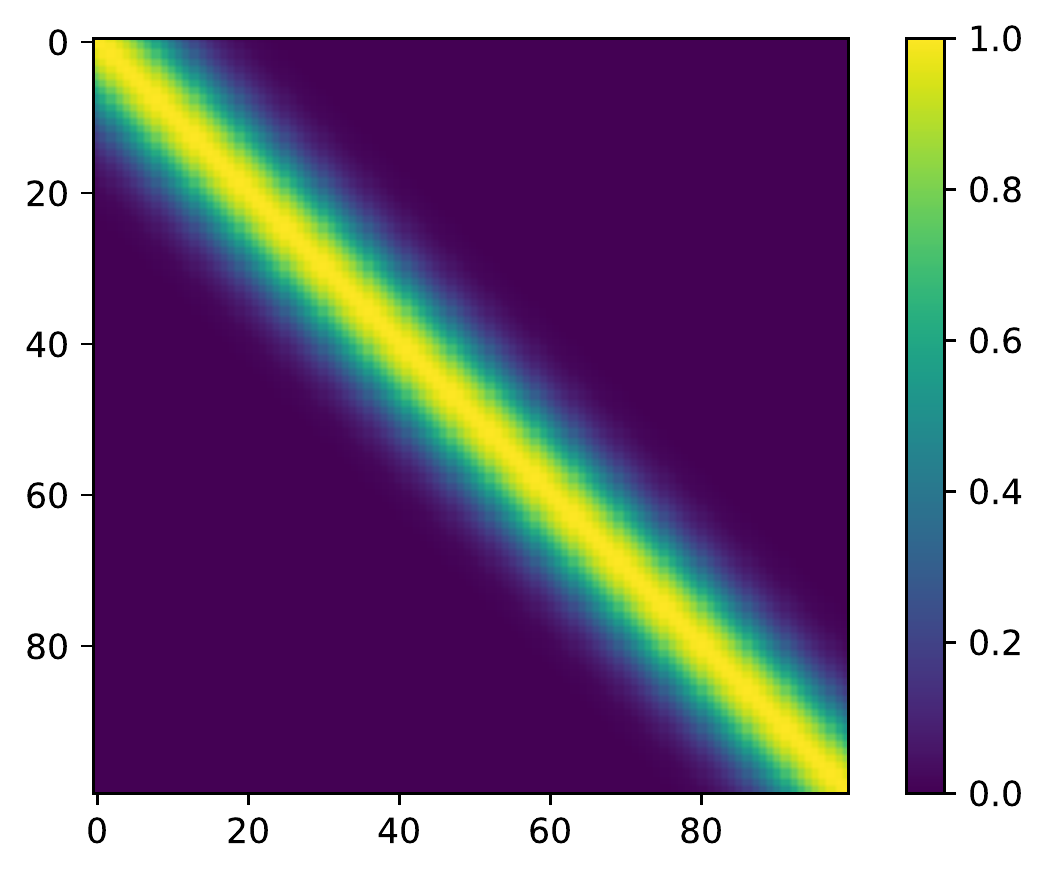}
        \caption{Visualisation of matrix $\mathbf{Q}$ ($l = 100$)}
        \label{fig: Q_100}
\end{figure}

In the third example, we use a highly correlated state noise process. All entries of matrix $\mathbf{Q}$ are strictly positive and expressed as
$Q(i,j) = \exp{(-(i - j)^{2} / 100)}$, i.e. correlation between state variables is higher as their indices are closer (see Fig. \ref{fig: Q_100} for visualisation). Using such correlated noise leads to a stronger correlation in the posterior distribution of the state process. For a number of blocks $K$ that is a divisor of $d_x$, an obvious choice for the partition consists in forming blocks of adjacent state variables: $B_k =\lbrace (k-1)\times \frac{d_x}{K}+1 ,.., k\times \frac{d_x}{K}, ~ \forall k \in \lbrace 1:K \rbrace$. This "known" partition in line with $\mathbf{Q}$ causes the fewest correlation breaks.


\begin{figure}[H]
    \centering
    \includegraphics[height = 4.5cm]{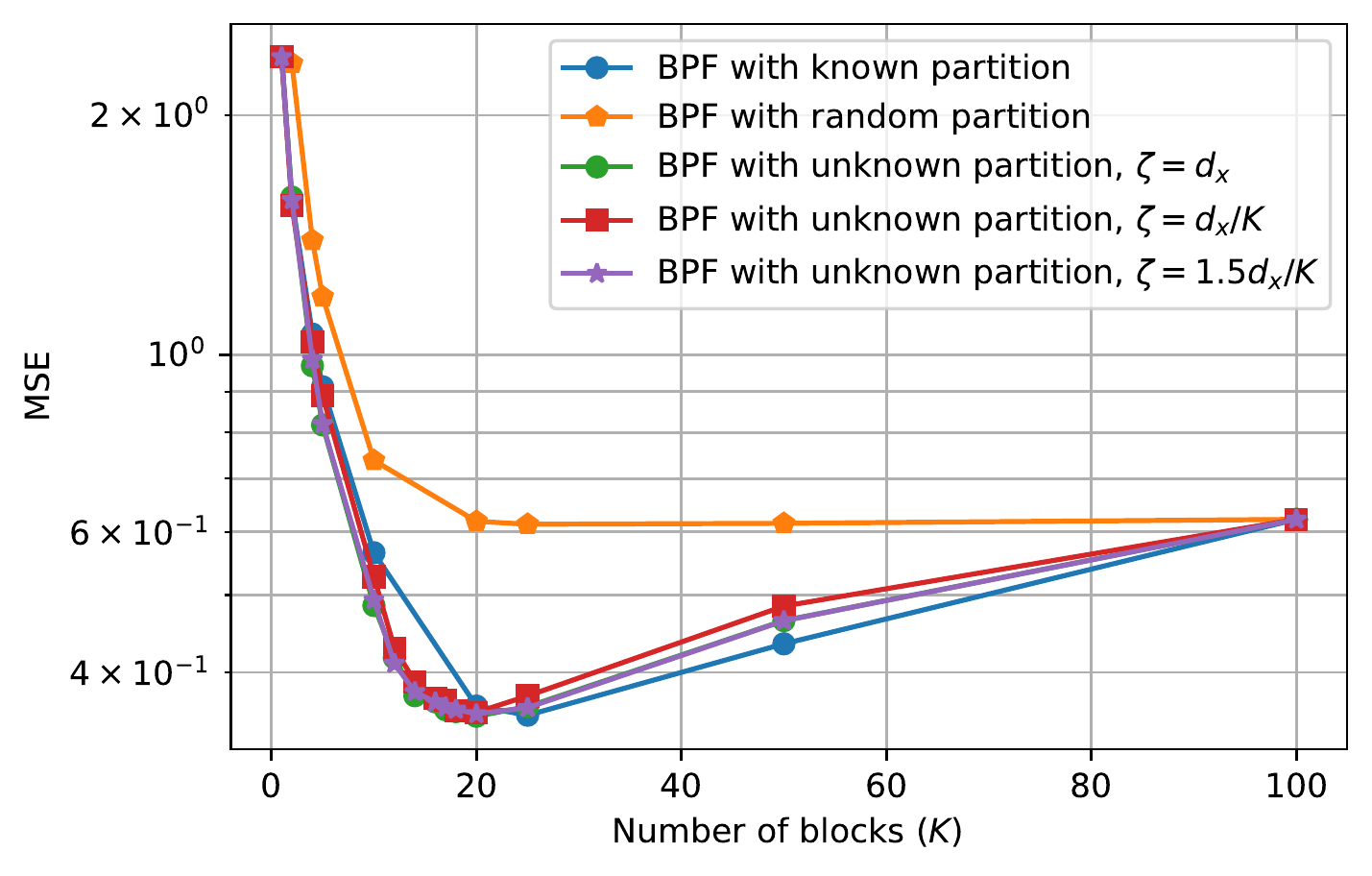}
    \caption{MSE performance of BPFs versus the number of blocks $K$ in a linear Gaussian model for $N_p=500$. 
    State noise with a dense correlation matrix.}
    \label{fig:lin_gauss_dense}
\end{figure}

Figure \ref{fig:lin_gauss_dense} depicts MSE performance of several versions of BPF versus the number of blocks $K$. For BPF with our partitioning method (case of unknown partition), several values of $\zeta$ are considered: $\zeta=d_x$ (unconstrained SC), $\zeta=\left\lceil  \frac{d_x}{K}\right\rceil$ (same-size block partition) and $\zeta=\left\lceil 1.5 \frac{d_x}{K}\right\rceil$ (more choice for the partition). The number of particles is chosen to be equal to $N_p = 500$.

For any number of blocks $K$, the PBF with the known partition picked from the structure of $\mathbf{Q}$ performs better than the BPF using a random partition that can group together remote state variables which are less correlated than adjacent ones. BPFs learning the partition on the fly with our SC method exhibit similar performance to the BPF with the known partition, which shows the ability of our approach to provide relevant partitions. Note that thanks to the flexibility of Algo.\ref{alg : SP Block particle filter}, an arbitrary number of blocks $K \in \lbrace1,..,d_x \rbrace$ can be requested in BPF with unkown partition while the known partition is only defined for a value of $K$ that is a divisor of $d_x$. Consequently, corresponding curves do not exhibit the same number of points.

For a value of $K$, we observe a low impact of the constraint $\zeta$ used in SC on the performance. In contrast, the figure shows that MSE depends on the number of blocks $K$ and thus on the block sizes. MSE reaches its lowest value for $K$ around $20$ in the range of $15$ to $25$. This range of optimal values corresponds to a bias / variance trade-off. If $K$ is smaller, the block sizes are bigger, BPF is prone again to the curse of dimensionality and the variance increases. If $K$ is larger, the block sizes are smaller, too much correlation across the state variables is ignored and the bias gets higher. It is interesting to compare the optimal values of $K$ with the approximate one given by the rule of thumb \cite{rebeschini2015can} given in subsection \ref{SectionBPFPerformance}. 
For $N_p=500$ particles, it recommends a block size $|B_k|$ about $\log N_p = 6.2$ and thus a number of blocks around $16.1$. Using this rule is a good way to fix the number of blocks to be given as input parameter to Algo. \ref{alg : SP Block particle filter}.

\subsection{Lorenz 96 non-linear model}

In this section, we move to a more general non-linear model for which only PF based solutions can be employed. 
We consider the Lorenz 96 model \cite{lorenz1998optimal} which is a low-order discrete chaotic model spanned by the following set of one-dimensional ordinary differential equations (ODEs):

\begin{equation}
    \frac{\mathrm{d} x_t(n)}{\mathrm{~d} t}=\left(x_t(n+1)-x_t(n-2)\right) x_t(n-1)-x_t(n)+F, \quad n=1 \ldots d_x
\end{equation}

where $x_t(n)$ is the $n^{\textrm{th}}$ state variable of $\mathbf{x}_t$ and indices follow periodic boundary conditions: $x_t(-1)=x_t(d_x-1)$, $x_t(0)=x_t(d_x)$ and $x_t(d_x+1)=x_t(1)$. These ODEs are integrated using a fourth-order Runge–Kutta method with a time step of a $0.05$ time unit. In the standard configuration, the system dimension is $d_x = 40$ and the forcing constant is $F = 8$, which
yields chaotic dynamics with a doubling time around $0.42$ time unit. The initial state vector is $\mathbf{x}_0 \sim \mathcal{N}(\mathbf{0}, 0.01 \times \mathbf{I})$. The observation system setting is the same as in \cite{bocquet2010beyond}. One of every two state variables is observed, so observations are expressed as: $y_t(n) = x_t(2n-1) + v_t(n)$ with $v_t(n) \sim \mathcal{N}(0, 1)$, for $n=1 \ldots d_x/2$. 

\begin{figure}[H]
     \centering
     \begin{subfigure}[b]{0.49\textwidth}
         \centering
         \includegraphics[width=\linewidth]{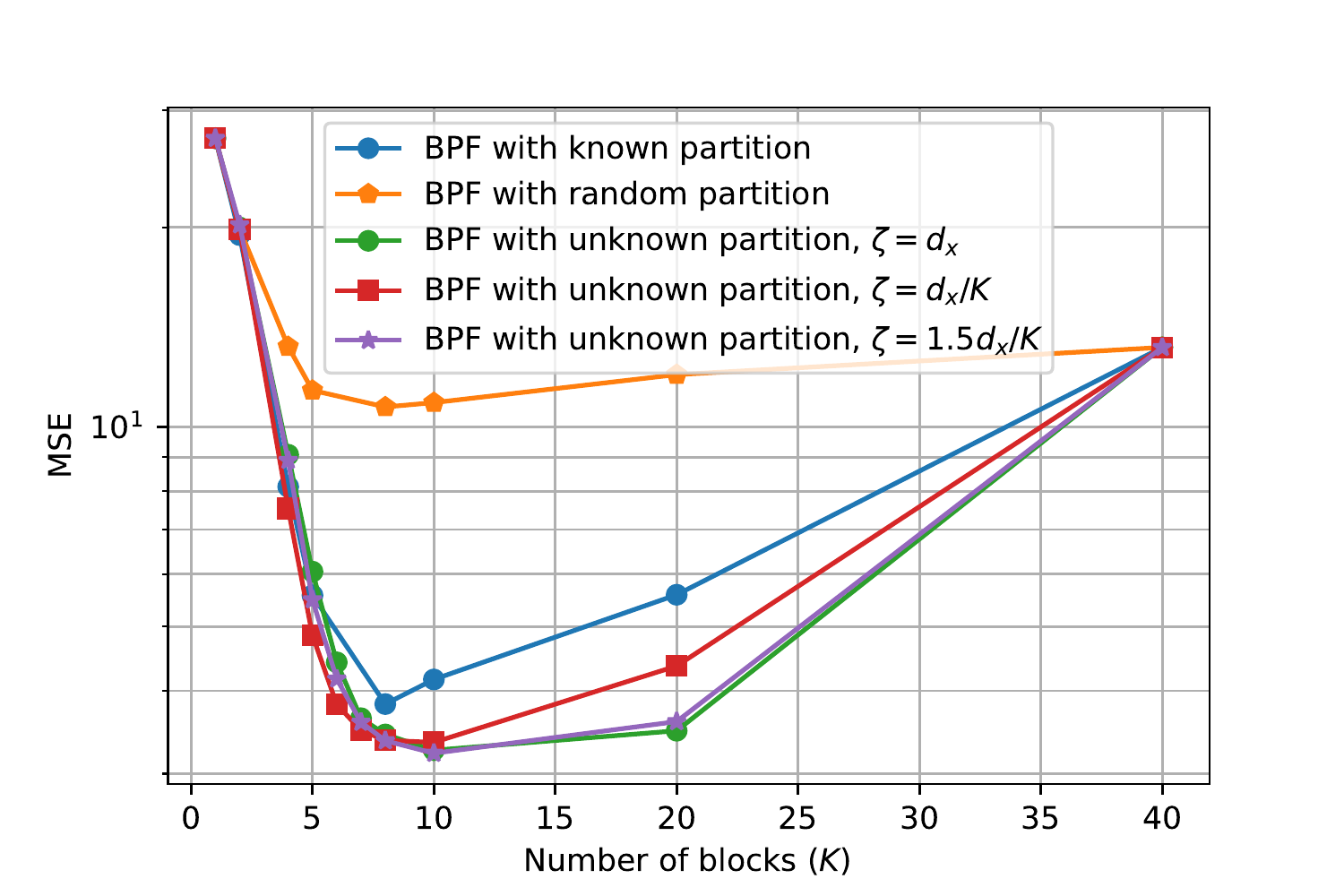}
         \caption{Independent state noise}
         \label{fig: inedpendent}
     \end{subfigure}
     \hfill
     \begin{subfigure}[b]{0.49\textwidth}
         \centering
         \includegraphics[width=\linewidth]{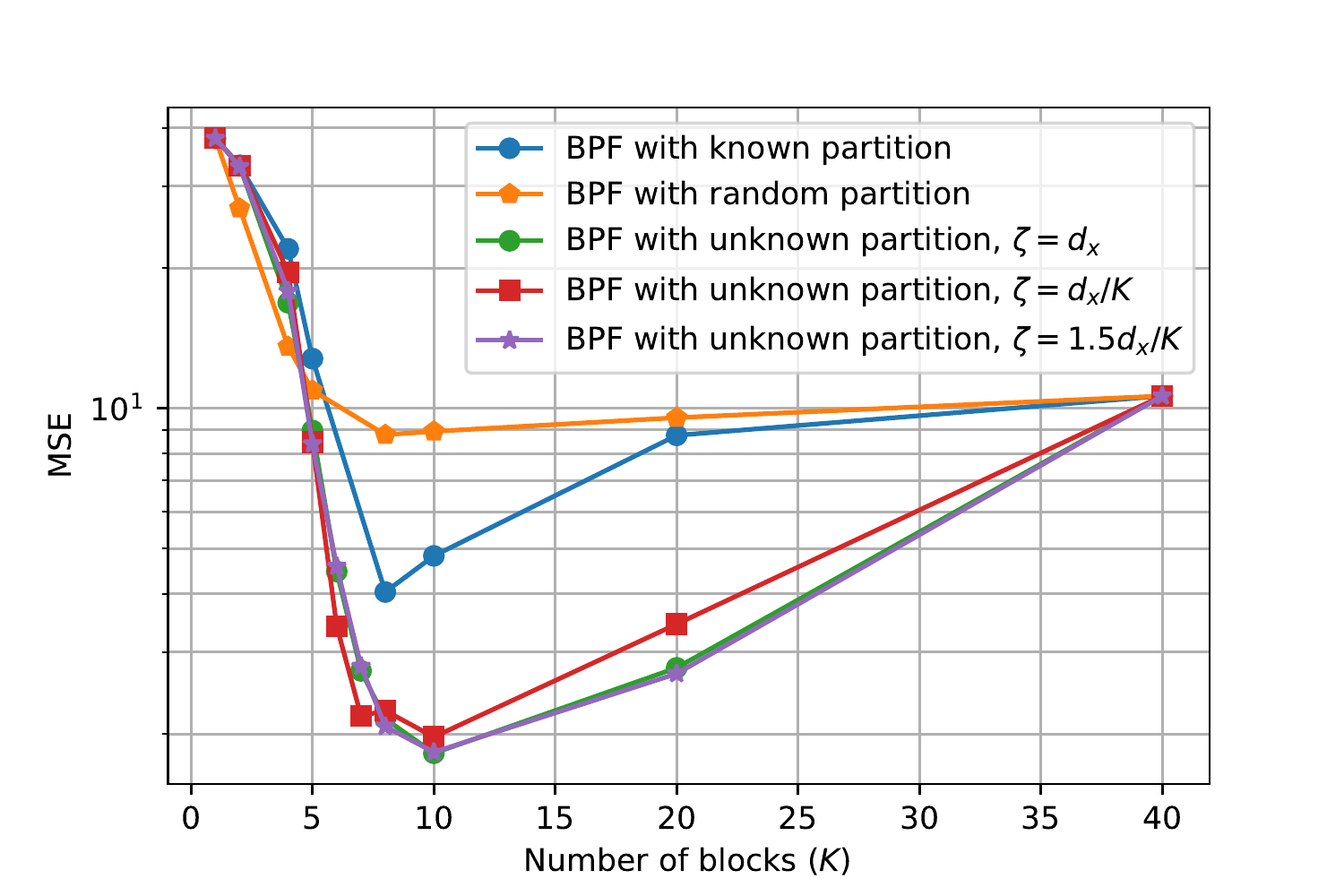}
         \caption{Correlated state noise}
         \label{fig: correlated}
     \end{subfigure}
        \caption{MSE performance of BPFs versus the number of blocks $K$ in a Lorenz 96 model for $N_p=1000$. }
        \label{fig:lorenz}
\end{figure}

Owing to the non-linearity of the state equation, we first perturb the state by an independent state noise. We then introduce a correlated state noise whose covariance matrix $\mathbf{Q}$ is the same as the one previously described in subsection \ref{sssec:densecov}. MSE performances of different instances of BPF versus the number of blocks $K$ are reported in Fig. \ref{fig:lorenz} for both types of state noise. These results are obtained from $200$ simulations of length $100$ time steps.

This non-linear model is more complex, so the number of particles is set at a larger value $N_p=1000$ in BPFs. 
It is also harder to deduce from the model a candidate partition and identify what is called the known partition. 
For a block number $K$ that is a divisor of $d_x$, a simple and natural choice consists in partitioning into blocks: $B_k =\lbrace (k-1)\times \frac{d_x}{K}+1 ,.., k\times \frac{d_x}{K}, ~ \forall k \in \lbrace 1:K \rbrace$ because by construction, the Lorenz 96 model induces correlation among consecutive state variables at each time step.  
For BPF with SC (case of unknown partition), several values of the constraint $\zeta$ are considered: $\zeta=d_x$ (unconstrained SC), $\zeta=\left\lceil  \frac{d_x}{K}\right\rceil$ (same-size block partition) and $\zeta=\left\lceil 1.5 \frac{d_x}{K}\right\rceil$. 


In both cases (independent and correlated state noise), the benefits of our partitioning approach based on constrained SC are highlighted. Algo.$\ref{alg : SP Block particle filter}$ performs significantly better whenever $K\geq 8$. For the Lorenz 96 model, the PBF with the known partition is less efficient, in particular with correlated state noise, which confirms that it is difficult to derive a good partition directly from the model. Among BPFs with partitions learnt on the fly, we also observe a moderate sensitivity w.r.t. the constraint $\zeta$.
The best performances, i.e. the most relevant partitions, are obtained when the maximal size constraint is looser ($\zeta=\left\lceil 1.5 \frac{d_x}{K}\right\rceil$) or omitted ($\zeta = d_x$). If a same-size block partition ($\zeta=\left\lceil \frac{d_x}{K}\right\rceil$) is enforced, MSE is slightly higher. 

As for the previous linear Gaussian model, MSE depends on the number of blocks $K$ and thus on the block sizes. MSE reaches its minimum value for $K$ around $10$, approximately in the range of $8$ to $10$. Smaller values of $K$ restore the curse of dimensionality and larger values of $K$ causes more correlation breaks. The rule of thumb \cite{rebeschini2015can} recommends a block size $|B_k|$ about $\log N_p = 6.9$ for $N_p=1000$ particles, and thus a number of blocks around $5.8$. Again, this approximate rule can help to determine the number of blocks to be given as input parameter to Algo.\ref{alg : SP Block particle filter}. We suggest to use a value a bit over the value given by this rule. 



\section{Conclusion}
\label{sec:conclusion}

In this paper, a new version of the block particle filtering algorithm is introduced. Compared to existing ones
, this version is able to detect automatically relevant state space partitions in order to perform the correction and resampling steps of traditional particle filtering in smaller dimensional subspaces thereby circumventing the notorious curse of dimensionality. This is made possible by running a spectral clustering algorithm using an estimate of the covariance matrix of the predictive posterior distribution as similarity matrix and by constraining the detected clusters to have sizes below a prescribed maximal threshold that prevents the resurgence of the curse of dimensionality.%

This new approach has been validated on various state space models including linear Gaussian and non-linear ones. In all cases, our algorithm is able to detect meaningful partitions for various numbers of blocks and achieves comparable or significantly better mean square error performances than the traditional block particle filter that relies on default partition choices motivated by the state noise correlation matrix structure or the state transition equation.


Although the rule of thumb \cite{rebeschini2015can} can give an order of magnitude for the block size given a number of particles, an open question that remains to be solved is how to exactly determine the optimal number of blocks.
A possible direction for future work consists in coupling the present contribution with the work in \cite{min2021parallel} that leverages a parallelisation scheme to optimize the deployment of the block particle filter. In particular, one could use filters with different numbers of blocks in parallel. This would at least make the approach less sensitive compared to a situation where a single number of blocks is used.

\bibliographystyle{plain}
\bibliography{mybibfile}

\end{document}